\documentclass{article} % For LaTeX2e
\usepackage{iclr2027_conference,times}
\usepackage{url}
\usepackage{amsmath,amssymb,amsthm}
\usepackage{tikz}
\usetikzlibrary{arrows.meta,positioning,calc,fit,backgrounds,%
  decorations.pathmorphing,shapes.geometric}
\definecolor{nrblue}{RGB}{40,101,145}
\definecolor{rorange}{RGB}{206,110,38}
\definecolor{slate}{RGB}{95,105,115}
\usepackage{booktabs}
\usepackage{graphicx}
\usepackage{xcolor}
\usepackage{multirow}
\usepackage{enumitem}
\usepackage[colorlinks=true,allcolors=blue]{hyperref}

\newcommand{\zi}{z_{\mathrm{nr}}}    %
\newcommand{\zn}{z_{r}}              %
\newcommand{\di}{d_{\mathrm{nr}}}
\newcommand{\dn}{d_{r}}
\newcommand{\Real}{\mathbb{R}}
\newcommand{\Exp}{\mathbb{E}}
\newcommand{\KL}{\mathrm{KL}}

\usepackage{colortbl}
\definecolor{mygray}{gray}{0.90}      %
\definecolor{mygray1}{gray}{0.96}     %
\newlength\savedwidth
\newcommand\thickhline{\noalign{\global\savedwidth\arrayrulewidth
  \global\arrayrulewidth 1pt}\hline\noalign{\global\arrayrulewidth\savedwidth}}

\newcounter{algo}
\newcommand{\algin}{\hspace*{1.1em}}
\newenvironment{algo}[1]{%
  \begin{figure}[t]\centering\small\refstepcounter{algo}%
  \begin{minipage}{\linewidth}
  \hrule height 0.9pt \vspace{2pt}
  \textbf{Algorithm \thealgo:} #1\par\vspace{2pt}\hrule height 0.4pt \vspace{3pt}
  \begin{list}{\scriptsize\arabic{algoline}:}{\usecounter{algoline}%
    \setlength{\leftmargin}{2.1em}\setlength{\itemsep}{1pt}\setlength{\parsep}{0pt}
    \setlength{\topsep}{0pt}\setlength{\labelsep}{0.5em}}%
}{%
  \end{list}\vspace{3pt}\hrule height 0.9pt
  \end{minipage}\end{figure}}
\newcounter{algoline}

\makeatletter
\renewenvironment{table}
  {\setlength{\abovecaptionskip}{0pt}\setlength{\belowcaptionskip}{7pt}\@float{table}}
  {\end@float}
\renewenvironment{table*}
  {\setlength{\abovecaptionskip}{0pt}\setlength{\belowcaptionskip}{7pt}\@dblfloat{table}}
  {\end@dblfloat}
\makeatother

\title{DRIFT: Disentangled Responsive-Invariant Flow Transport for Single-Cell
Perturbation Prediction}
\author{Mustapha Bounoua\thanks{Correspondence to: \texttt{mustapha.bounoua@eurecom.fr}},
Giulio Franzese \& Pietro Michiardi \\
Department of Data Science \\
EURECOM, France}
\iclrfinalcopy
\begin{document}
\maketitle

\begin{abstract}
Predicting cellular responses to perturbations is a central problem in cellular biology, with broad applications in systems biology and drug discovery.
This task is challenging because cellular responses can be complex and cell-state dependent, intrinsic cell-to-cell variability can be confounded with perturbation effects, and destructive single-cell RNA sequencing precludes paired measurements of the same cell before and after treatment.
Flow matching transports control cells to perturbed states flexibly, but acting on the full cell state can confound perturbation effects with pre-existing cell-to-cell variability.
Disentangled approaches separate responsive from invariant components, but model perturbations through prescribed mechanisms, such as latent shifts or graph edits, limiting their flexibility.
We address both limitations in a unified framework.
A variational encoder disentangles each cell into an invariant block, capturing state unaffected by the perturbation, and a responsive block, capturing state it changes, through conditional priors and an information-theoretic invariance constraint.
Conditional flow matching transports only the responsive block, conditioned on the perturbation and invariant state, yielding a flexible, data-driven model of perturbation effects without confounding pre-existing variability.
Across several benchmarks, our method outperforms the strongest published method in settings involving combinatorial and unseen perturbation prediction.

\end{abstract}

\section{Introduction}
\label{sec:intro}

A perturbation experiment measures what a cell does when a gene is activated or silenced, or when a drug is applied.
In a pooled screen \citep{dixit2016perturb,replogle2020combinatorial,replogle2022mapping,srivatsan2020sciplex}, a large population of cells is treated at once, each cell receives a single perturbation identified by a sequencing barcode, and the transcriptome of every cell is measured afterwards.
The result is two kinds of data: untreated control cells, and perturbed cells grouped by which perturbation treatment they received.
The prediction task is to predict the distribution of transcriptomes that control cells would exhibit under unseen or combinatorial perturbations.

Such a model would let researchers predict, in silico, how cells respond to perturbations they have never been exposed to \citep{bunne2023cellot,lotfollahi2023cpa}. This would allow researchers to screen perturbations computationally before applying them in vitro, prioritize combinations in spaces too large to explore experimentally, and estimate the effect of treatments for which no experiment exists. These capabilities are important for drug discovery, precision medicine, and cell engineering \citep{bunne2023cellot,lotfollahi2023cpa}.

These screens have made training data abundant, but the measurement itself withholds the very quantity a per-cell model requires.
Sequencing destroys the cell it measures ~\citep{kester2018single, chen2019high, schiebinger2019optimal,demetci2022scot} which results in unpaired control and treatment populations which prohibit the measurement of a per-cell causal effect and the outcome of cell under treatment is never measured. Therefore, every approach in the field relies on assumptions to bridge the gap between the observed data and the quantity of interest.

Existing approaches close this gap with different modeling choices.
Statistical methods predict the response with simple estimators, such as the control mean or a linear map, and carefully tuned linear baselines are competitive with respect to far more elaborate models \citep{ahlmann2025deep}.
Foundation models pretrained on large unperturbed corpora \citep{cui2024scgpt,theodoris2023geneformer,rosen2023uce,yang2022scbert,hao2024large,yang2024genecompass,zeng2025cellfm,khan2023reusability} supply broad representations but are repeatedly found to miss perturbation-specific effects \citep{ahlmann2025deep,csendes2025benchmarking,kedzierska2025zeroshot}, which has prompted more discriminating evaluation protocols \citep{vinas2025systema,roohani2025virtualcell} that we adopt.
Graph-based models reach unseen perturbations by placing genes in a structure taken from external biology: a gene co-expression or Gene Ontology graph \citep{roohani2024gears,kamimoto2023dissecting,wu2023graphvci}, or gene modules shared across perturbations \citep{scbig2026}.
Generative models learn the distribution of the whole perturbed cell state, in expression space or in a latent space \citep{adduri2025predicting,yu2026scdfm}.
Transport methods \citep{schiebinger2019optimal,bunne2023cellot,klein2023genot,klein2025cellflow,scbig2026} model the shift of an entire distribution from control to perturbed.
Disentangled and causal latent-variable models \citep{lotfollahi2023cpa,lotfollahi2019scgen,weinberger2023isolating,%
yu2025perturbnet,xing2025gperturb,bereket2023samsvae,lopez2023learning,%
perturbedvae2026} instead isolate a latent axis along which a perturbation acts.
This factorization is intended to enable extrapolation to unobserved perturbation
combinations \citep{lotfollahi2023cpa,bereket2023samsvae}, generalization to held-out cell
states \citep{aliee2023invae,zhang2024scdisinfact}, and a causal interpretation of the
perturbation factor \citep{lopez2023learning,gao2025causal}.
Taken together, these methods either model the full cell state or impose a disentanglement scheme with a preselected, potentially inflexible perturbation mechanism.

We focus on two complementary approaches. 
Transport methods assume minimal displacement between control and perturbed cells and learn the transport map from data, but move the whole cell representation, either in a latent space or in the original expression space \citep{bunne2023cellot,klein2025cellflow,yu2026scdfm}.
The invariant part of the cell then moves together with the responsive part, and the perturbation is confounded with the cell characteristics that are invariant to treatment. Disentanglement based models separate these invariant characteristics from the responsive state, but they generally assume a causal mechanism for how the perturbation acts: an additive vector \citep{lotfollahi2023cpa}, a sparse mechanism shift \citep{lachapelle2022disentanglement,bereket2023samsvae}, or an edit to a latent causal graph \citep{an2025sccausalvi}.
These 
approaches commit to an inductive prior about perturbation dynamics and thus do not generalize to incompatible biological programs. 

Our proposed method, which we call \textsc{Drift} (Disentangled Responsive-Invariant Flow Transport), separates the part of a cell that is responsive to the perturbation from the part that is invariant \citep{higgins2018definition,winter2022unsupervised,garrido2023sie}, and then models the perturbation mechanism as a per-cell transport that moves only the responsive part, leaving fixed the part carrying cell identity and the other perturbation-invariant factors. The disentanglement is learned following the auxiliary-variable method \citep{khemakhem2020ivae} and related lines of work \citep{khemakhem2020icebeem,sorrenson2020disentanglement,hyvarinen2016tcl,lu2021nonlinear,aliee2023invae,vonkugelgen2021self}, augmented with an information-theoretic constraint of invariance.
The dynamics of the perturbation is modeled through a conditional flow \citep{lipman2023flow,tong2024cfm,klein2025cellflow} that moves the variant block while keeping the invariant part fixed.

We summarize our contributions as follows. 
1) We propose an information-theoretic disentanglement scheme that uses auxiliary-variable conditional priors together with a mutual-information-based invariance constraint to split the latent space of a cell into an invariant block, insensitive to the perturbation, and a variant block, that captures the perturbation-responsive part. 
2) We use a conditional flow matching model, conditioned jointly on the perturbation and on the invariant latent of the cell, that transports the responsive (variant) part of the latent representation, leaving the invariant part fixed. 
3) We obtain state-of-the-art results across multiple benchmarks: our method outperforms the strongest published method on every dataset we experimented with.

\section{Method}
\label{sec:method}

\newcommand{\celldraw}[3]{%
  \begin{scope}[shift={#1}]
    \draw[draw=#2!55, fill=#2!7, line width=0.8pt]
      plot[smooth cycle, tension=0.9] coordinates
      {(0.58,0.05)(0.34,0.40)(0.00,0.54)(-0.37,0.41)(-0.55,0.03)%
       (-0.39,-0.37)(-0.02,-0.54)(0.39,-0.36)};
    \draw[draw=#2!50, fill=#2!20] (0.05,0.02) ellipse (0.20 and 0.16);
    \fill[#2!60] (0.09,0.05) circle (0.042);
    \fill[#2!45] (-0.26,0.20) circle (0.042);
    \fill[#2!45] (0.28,-0.23) circle (0.047);
    \ifnum#3=1 \fill[#2!45] (-0.19,-0.25) circle (0.036);\fi
  \end{scope}}
\newcommand{\boltbadge}[1]{\begin{scope}[shift={#1}]
  \fill[white,draw=orange!65!black,line width=0.4pt] (0,0) circle (0.16);
  \fill[yellow!85!orange,draw=orange!80!black,line width=0.2pt]
    (0.03,0.105)--(-0.048,-0.008)--(0.014,-0.008)--(-0.032,-0.112)%
    --(0.056,0.014)--(0.0,0.014)--cycle;
\end{scope}}
\newcommand{\flake}[1]{\begin{scope}[shift={#1},nrblue!75,line width=0.4pt]
  \foreach \a in {30,90,150}{\draw (\a:0.10)--(\a+180:0.10);}
  \foreach \a in {30,90,150,210,270,330}{%
    \draw (\a:0.10)--++(\a+150:0.035); \draw (\a:0.10)--++(\a-150:0.035);}
\end{scope}}
\newcommand{\gear}[1]{\begin{scope}[shift={#1},draw=rorange!80!black,line width=0.7pt]
  \foreach \a in {0,45,90,135,180,225,270,315}{\draw (\a:0.052)--(\a:0.093);}
  \draw[fill=rorange!18] (0,0) circle (0.052);
  \fill[white] (0,0) circle (0.019);
\end{scope}}
\newcommand{\gauss}[2]{%
  \begin{scope}[shift={#1}]
    \draw[#2!55!black, line width=0.5pt] (-0.36,0)--(0.36,0);
    \draw[#2, line width=0.9pt] plot[domain=-0.34:0.34,samples=24]
      (\x,{0.27*exp(-(\x*\x)/0.024)});
  \end{scope}}
\newcommand{\cloud}[2]{%
  \begin{scope}[shift={#1}]
    \fill[#2!10] (0,0) ellipse (0.39 and 0.31);
    \foreach \p in {(0,0.02),(0.18,0.11),(-0.15,0.09),(0.10,-0.13),%
                    (-0.12,-0.10),(0.23,-0.03),(-0.23,-0.02),(0.03,0.18),(-0.05,-0.18)}
      \fill[#2!75] \p circle (1.4pt);
  \end{scope}}
\newcommand{\encoder}[1]{\filldraw[fill=nrblue!6,draw=nrblue!55,line width=0.8pt,%
  shift={#1}] (-0.58,0.58)--(0.52,0.4)--(0.52,-0.4)--(-0.58,-0.58)--cycle;}
\newcommand{\decoder}[1]{\filldraw[fill=nrblue!6,draw=nrblue!55,line width=0.8pt,%
  shift={#1}] (-0.52,0.4)--(0.58,0.58)--(0.58,-0.58)--(-0.52,-0.4)--cycle;}

\colorlet{zmid}{rorange!50!slate}   %
\begin{figure}[t]
\centering
\resizebox{\textwidth}{!}{%
\begin{tikzpicture}[>={Stealth[length=4pt,width=3.2pt]}, line cap=round,
  font=\small,
  chip/.style={rounded corners=2.5pt, draw=nrblue, fill=nrblue!12, line width=1pt,
               minimum width=11mm, minimum height=7mm},
  fchip/.style={rounded corners=5.2pt, draw=zmid!80, fill=zmid!8, line width=0.9pt,
                inner xsep=10pt, inner ysep=5.5pt},
  cat/.style={rounded corners=2pt, draw=slate!55, fill=slate!7, line width=0.8pt,
              inner sep=2.5pt, font=\footnotesize},
  klab/.style={font=\scriptsize, text=slate!85},
  var/.style={circle, draw=slate!55, fill=slate!7, inner sep=0pt,
              minimum size=5.5mm, font=\footnotesize},
  wave/.style={rorange!85!black, line width=1.1pt,
               decorate, decoration={snake, amplitude=0.5pt, segment length=6pt,
               post length=3.5pt}},
  keep/.style={nrblue, line width=1.1pt}]

\begin{scope}
  \celldraw{(0,0.72)}{nrblue}{0}   \node[klab] at (0,1.38) {Control};
  \celldraw{(0,-0.72)}{rorange}{1}\boltbadge{(0.41,-0.28)}
  \node[klab] at (0,-1.38) {Perturbed};
  \encoder{(2.1,0)} \node at (2.1,-0.12) {$q_\phi$}; \gear{(2.1,0.3)}
  \draw[->,black!65] (0.55,0.62) to[out=0,in=160] (1.5,0.28);
  \draw[->,black!65] (0.55,-0.62) to[out=0,in=-160] (1.5,-0.28);

  \node[chip] (znr) at (4.2,0.8) {$\zi$};
  \cloud{(4.2,-0.8)}{slate}
  \node[slate!80!black,font=\small] at (4.72,-0.56) {$\zn$};
  \draw[->,black!65] (2.66,0.2) to[out=18,in=180]  (znr.west);
  \draw[->,black!65] (2.66,-0.2) to[out=-18,in=180] (3.8,-0.8);

  \node[var] (c) at (2.55,1.9) {$c$};
  \gauss{(4.2,1.9)}{nrblue} \gear{(4.62,2.02)}
  \node[klab,anchor=west] at (4.82,1.9) {$p(\zi\mid c)$};
  \draw[->,slate!70] (c) -- (3.78,1.9);
  \draw[->,nrblue!70,dashed] (4.2,1.74) -- (znr.north);
  \node[var] (u) at (2.55,-1.9) {$u$};
  \gauss{(4.2,-1.9)}{rorange} \gear{(4.62,-1.78)}
  \node[klab,anchor=west] at (4.82,-1.9) {$p(\zn\mid u)$};
  \draw[->,slate!70] (u) -- (3.78,-1.9);
  \draw[->,rorange!75,dashed] (4.2,-1.74) -- (4.2,-1.12);

  \node[font=\scriptsize,text=black!70] at (5.15,0.92) {$\oslash$};
  \node[klab,anchor=west,align=left] at (5.33,0.92)
    {Invariance\\[-2pt]$I(\zi;u){=}0$};

  \node[cat] (cat1) at (8.35,0) {$[\zi,\zn]$};
  \draw[->,black!65] (znr.east) to[out=-12,in=150] (cat1.north west);
  \draw[->,black!65] (4.6,-0.82) to[out=-2,in=210] (cat1.south west);
  \decoder{(9.6,0)} \node at (9.6,-0.12) {$p_\theta$}; \gear{(9.6,0.3)}
  \draw[->,black!65] (cat1.east) -- (9.0,0);
  \celldraw{(11.5,0.62)}{nrblue}{0}
  \celldraw{(11.5,-0.62)}{rorange}{1}\boltbadge{(11.9,-0.2)}
  \draw[->,black!65] (10.2,0.12) to[out=15,in=200] (10.98,0.55);
  \draw[->,black!65] (10.2,-0.12) to[out=-15,in=160] (10.98,-0.55);
  \node[klab,align=center] at (11.5,1.36) {Reconstruction $\hat{x}$};
\end{scope}

\begin{scope}[yshift=-49mm]
  \celldraw{(0,0)}{nrblue}{0}   \node[klab] at (0,-0.96) {Control cell};
  \encoder{(2.1,0)} \node at (2.1,-0.12) {$q_\phi$}; \flake{(2.1,0.3)}
  \draw[->,slate!70] (0.58,0) -- (1.5,0);

  \node[chip] (znr2) at (4.2,0.85) {$\zi$};
  \cloud{(4.2,-0.85)}{slate}
  \draw[->,slate!70] (2.66,0.22) to[out=20,in=180] (znr2.west);
  \draw[->,slate!70] (2.66,-0.22) to[out=-20,in=180] (3.78,-0.85);

  \cloud{(5.95,-0.85)}{zmid}
  \cloud{(7.55,-0.85)}{rorange}
  \draw[wave,->] (4.62,-0.85) -- (5.50,-0.85);
  \draw[wave,->] (6.40,-0.85) -- (7.12,-0.85);
  \node[klab] at (4.20,-1.32) {$t{=}0$};
  \node[klab] at (7.55,-1.32) {$t{=}1$};

  \node[fchip] (vfield) at (6.35,0.10) {\hspace{3.1mm}$v_\theta(z_t,t\mid \zi,\,u)$};
  \gear{($(vfield.west)+(0.34,0)$)}
  \draw[zmid!65,dashed,line width=0.7pt,->] ($(vfield.south)+(-0.4,0)$) -- (5.95,-0.50);

  \node[var] (u2) at (5.95,-1.72) {$u$};
  \draw[->,slate!70] (u2.north) -- (5.95,-1.18);

  \node[cat] (cat2) at (8.75,0) {$[\zi,\zn']$};
  \draw[keep,->] (znr2.east) to[out=16,in=172] (5.30,1.16) -- (8.15,1.16)
        to[out=0,in=96] (cat2.north);
  \draw[->,rorange!75] (8.05,-0.85) to[out=18,in=250] (cat2.south);
  \decoder{(9.9,0)} \node at (9.9,-0.12) {$p_\theta$}; \flake{(9.9,0.3)}
  \draw[->,slate!70] (cat2.east) -- (9.35,0);
  \celldraw{(11.7,0)}{rorange}{1}\boltbadge{(12.1,0.4)}
  \draw[->,slate!70] (10.5,0) -- (11.15,0);
  \node[klab,align=center] at (11.7,-0.96) {Predicted\\perturbed cell};
\end{scope}

\begin{scope}[on background layer]
  \draw[draw=black!12, line width=0.5pt] (-1.0,-2.72) -- (12.75,-2.72);
\end{scope}
\node[font=\footnotesize\itshape, nrblue!55!black, anchor=west]  at (-0.95,2.66)
  {Stage 1 $\cdot$ Disentangle};
\node[font=\footnotesize\itshape, rorange!55!black, anchor=west] at (-0.95,-3.28)
  {Stage 2 $\cdot$ Transport};
\gear{(8.5,2.62)}   \node[klab,anchor=west] at (8.66,2.62) {Learned};
\flake{(10.35,2.62)}\node[klab,anchor=west] at (10.52,2.62) {Frozen};

\end{tikzpicture}}
\caption{\textbf{Overview of \textsc{Drift}.} The two stages share one architecture.
\emph{Stage~1} trains a variational autoencoder on control and perturbed cells, encoding each into an \emph{invariant} code
$\zi$ (driven by the covariates $c$: cell line, batch, sequencing depth) and a
\emph{responsive} latent $\zn$ (driven by the perturbation $u$). Each block is associated 
to a \emph{learned} conditional prior, $p(\zi\mid c)$ and $p(\zn\mid u)$, and $\zi$ is
optimized to carry no perturbation information, $I(\zi;u){=}0$; the concatenation
$[\zi,\zn]$ is decoded to reconstruct the input, control or perturbed. \emph{Stage~2}
freezes the encoder and decoder and learns the mechanism $p(\zn\mid\zi,u)$ as a
transport: a conditional flow $v_\theta(z_t,t\mid\zi,u)$ generates only the responsive
latent $\zn'$ from its control to its perturbed state along the interpolant $z_t$
($t{:}\,0\to1$), while $\zi$ is copied from the input cell. Decoding $[\zi,\zn']$
yields the predicted perturbed cell.}
\label{fig:overview}
\end{figure}
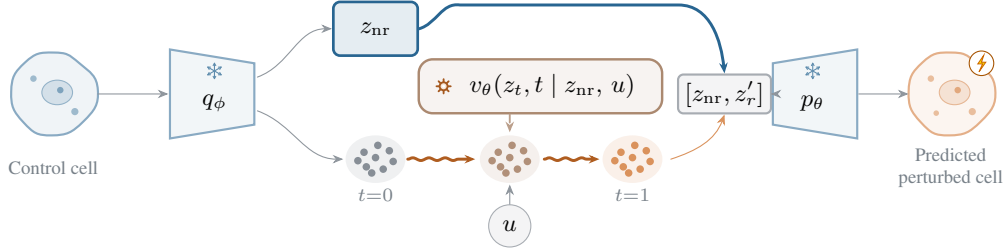

\textsc{Drift} models the perturbation effect as a mechanism that changes part of a cell's state and leaves the remaining part unchanged.
A cell is organized into an invariant block, holding the state a perturbation leaves untouched, and a variant (responsive) block, holding the state it changes, and the perturbation acts as a transport of the variant block alone.
Our method unfolds in two stages (Figure~\ref{fig:overview}).
Stage~1 (\S\ref{sec:genmodel} and \S\ref{sec:split}) learns a disentangled representation: an encoder maps a cell to the two blocks, each block is tied to a prior that depends on a different observed variable, and an information-theoretic constraint prevents perturbation information to leak into the invariant block.
Stage~2 (\S\ref{sec:flow}) learns how a perturbation moves the responsive block, as a conditional flow.
Inference (\S\ref{sec:inference}) combines the invariant state of an observed control cell with a response generated by the flow.

\subsection{Problem statement and generative model}
\label{sec:genmodel}

We observe single-cell RNA sequencing counts and write $x\in\Real^{G}$ for the expression profile of one cell over $G$ genes.
A perturbation is the experimental intervention applied to that cell: which gene is activated or silenced by CRISPR \citep{jinek2012programmable,gilbert2014genome}, or which chemical compound is added.
We label a perturbation by $u$, with $u=\varnothing$ for an untreated control.
The perturbation acts on the cell by changing its transcriptome, so it changes the distribution of $x$.

We also observe $c\in\Real^{K}$, the factors that vary across cells and are not caused by the perturbation, such as the cell line, and technical properties of the measurement.

We are given unpaired populations of control cells and of cells under various $u$; $c$ is observed per cell and the two populations are not assumed to share it.
The task is to predict, for a target perturbation $u^\star$ that may be absent from training, the distribution of profiles that observed control cells from the same dataset would take under $u^\star$.

We consider a latent state that splits into two blocks,
\begin{equation}
z=(\zi,\zn)\in\Real^{\di}\times\Real^{\dn},
\end{equation}
where $\zi$ is the \emph{invariant} (non-responsive) block, holding the state a perturbation leaves untouched, and $\zn$ is the \emph{variant} (responsive) block, holding the state a perturbation moves.

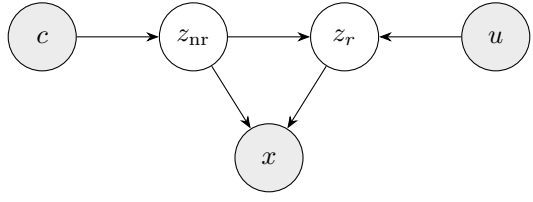
\begin{figure}[t]
\centering
\begin{tikzpicture}[>=Stealth,
  obs/.style={circle,draw,fill=gray!15,minimum size=9mm,inner sep=1pt},
  lat/.style={circle,draw,minimum size=9mm,inner sep=1pt}]
  \node[obs] (c)  at (0,0)    {$c$};
  \node[lat] (zi) at (2,0)    {$\zi$};
  \node[lat] (zn) at (4,0)    {$\zn$};
  \node[obs] (u)  at (6,0)    {$u$};
  \node[obs] (x)  at (3,-1.6) {$x$};
  \draw[->] (c)  -- (zi);
  \draw[->] (zi) -- (zn);
  \draw[->] (u)  -- (zn);
  \draw[->] (zi) -- (x);
  \draw[->] (zn) -- (x);
\end{tikzpicture}
\caption{The generative model of \textsc{Drift}; shaded nodes are observed. The
non-responsive block $\zi$ is generated from the assay covariates $c$; the responsive
block $\zn$ from the perturbation $u$ and $\zi$ (the mechanism); the profile $x$ from
both. The non-responsive block is invariant to the perturbation, $\zi\perp u$
(Eq.~\ref{eq:invariance}), the property that defines it, while the mechanism
$\zi\!\to\!\zn$ lets the response depend on the basal state.}
\label{fig:pgm}
\end{figure}

A decoder $p_\theta(x\mid\zi,\zn)$ maps the two blocks back to an expression profile.
The generative model is the chain of Fig.~\ref{fig:pgm},
\begin{equation}
p(x,\zi,\zn\mid u,c)
= \underbrace{p_\theta(x\mid \zi,\zn)}_{\text{decoder}}\;
\underbrace{p(\zn\mid \zi,u)}_{\text{mechanism}}\;
\underbrace{p(\zi\mid c)}_{\text{invariant state}} ,
\label{eq:chain}
\end{equation}
with edges $\{c\!\to\!\zi,\;\zi\!\to\!\zn,\;u\!\to\!\zn,\;\zi\!\to\!x,\;\zn\!\to\!x\}$ and $u\perp c$.
The invariant block is generated from the perturbation-independent factors $c$; the responsive block from the perturbation and from the invariant block of the same cell; and the expression profile from both blocks.
In our model, we encourage the invariant block to be independent of the perturbation,
\begin{equation}
\zi\perp u,
\label{eq:invariance}
\end{equation}
through a mutual-information penalty (\S\ref{sec:split}).
The mechanism $p(\zn\mid\zi,u)$ keeps $\zn$ dependent on $\zi$, which lets the response vary with the basal state.
Stage~1 (\S\ref{sec:split}) learns the two blocks and enforces \eqref{eq:invariance}; stage~2 (\S\ref{sec:flow}) learns the mechanism as a transport.

\subsection{Stage 1: disentangling the variant and invariant blocks}
\label{sec:split}

Disentanglement learned without supervision or inductive bias does not reliably recover meaningful factors \citep{locatello2019challenging,hyvarinen1999nonlinear}, which motivates tying each block to an observed variable.
In the causal representation learning view \citep{scholkopf2021toward}, a perturbation is a soft intervention: it changes the mechanism that generates a latent variable without fixing its value \citep{zhang2023identifiability}.
Following the auxiliary-variable construction \citep{khemakhem2020ivae, aliee2023invae}, each block is given a prior that depends on one observed variable only, which we assume to be Gaussian whose mean and (diagonal) covariance are functions of that variable,
\begin{equation}
p(\zi\mid c)=\mathcal N\!\big(\mu_{\mathrm{nr}}(c),\,\operatorname{diag}\sigma_{\mathrm{nr}}^2(c)\big),
\qquad
p(\zn\mid u)=\mathcal N\!\big(\mu_{r}(u),\,\operatorname{diag}\sigma_{r}^2(u)\big),
\label{eq:priors}
\end{equation}
that is, they are members of the conditional exponential family of \citet{khemakhem2020ivae} with Gaussian sufficient statistics. The priors are learned jointly with the encoder and decoder. The encoder is also fed with the auxiliary variables $u$ and $c$.

The encoder maps a cell to Gaussian posteriors over the two blocks conditioned on the cell and its auxiliaries, $q_\phi(\zi\mid x,u,c)$ and $q_\phi(\zn\mid x,u,c)$, following \citet{aliee2023invae}, and the decoder reconstructs $x$ from both blocks.
Writing $q_\phi$ for the joint posterior, we maximize the ELBO
\begin{equation}
\begin{aligned}
\mathcal{L}(\phi,\theta)=
\Exp_{q_\phi}\!\big[\log p_\theta(x\mid\zi,\zn)\big]
-\beta_t\Big(&
\KL\!\big[q_\phi(\zi\mid x,u,c)\,\|\,p(\zi\mid c)\big]\\
&+\KL\!\big[q_\phi(\zn\mid x,u,c)\,\|\,p(\zn\mid u)\big]\Big),
\end{aligned}
\label{eq:elbo}
\end{equation}
subject to the invariance constraint below.
Stage~1 regularizes $\zn$ toward the marginal prior $p(\zn\mid u)$ instead of the mechanism $p(\zn\mid\zi,u)$ of \eqref{eq:chain}.
The dependence of the response on the basal state is deferred to the stage-2 flow.

\noindent \textbf{Enforcing invariance.}
The conditional prior anchors each block to its auxiliary, but reconstruction does not prevent the invariant block from also carrying the perturbation.
We therefore encourage \eqref{eq:invariance} by penalizing mutual information between $\zi$ and $u$ \citep{moyer2018invariant},
\begin{equation}
\min_{\phi}\; I(\zi;u),
\label{eq:mi}
\end{equation}
using the contrastive log-ratio upper bound (CLUB) of \citet{cheng2020club},
\begin{equation}
\mathcal{L}\;\leftarrow\;\mathcal{L}\;-\;\lambda\, I_{\mathrm{CLUB}}(\zi;u).
\label{eq:club}
\end{equation}
This penalty encourages perturbation-dependent factors to concentrate in the responsive block, helping make $\zn$ the coordinate transported in stage~2.

\noindent \textbf{Regularizing the perturbation encoder.}

The perturbation identity $u$ is fed to the model through an embedded representation $e_u$, produced by a network that maps pretrained features of the targeted gene to a vector. Observed perturbations, available at training time, are paired with their corresponding perturbed cell population, whereas unseen perturbations have no example of the resulting outcome. Since the model must generalize to unseen genes, $e_u$ alone determines its behavior in the out-of-distribution regime, which requires the representation to be well structured so that unseen perturbations are meaningfully embedded and predictive of their effects. To this end, we regularize $e_u$ with three complementary terms. A linear head predicts the mean measured response of each training condition from its code, tying the embedding to the response it must explain. A second term constrains pairwise distances between codes to follow distances between the corresponding mean measured responses, so that perturbations with similar effects receive similar codes. Finally, Gaussian noise injected into the embedding during training prevents the network from collapsing into a lookup table of training conditions. These terms are auxiliary to the main objective in \eqref{eq:elbo} and are formally defined in Appendix~\ref{app:condpath}.

\noindent \textbf{Training.}
The CLUB bound is evaluated with an auxiliary predictor $q_\psi(u\mid \zi)$, trained to keep the bound tight while the encoder minimizes the bound $I_{\mathrm{CLUB}}(\zi;u)$. Two collapse modes must be guarded against.
First, the invariant block's $\KL$ term can collapse the posterior onto the prior; we prevent this with a $\KL$-annealing schedule, $\beta_t=\min(1,t/T_{\mathrm w})\,\beta$ \citep{higgins2017betavae}, which lets reconstruction dominate early so the block learns a useful representation before its $\KL$ cost is applied. Second, CLUB \citep{cheng2020club} is a valid upper bound only once $q_\psi$ has fit the current encoder; applying it from initialization drives $\zi$ to a trivial solution instead of an invariant one.
We therefore ramp the weight of \eqref{eq:club} up from zero over the same warm-up, while $q_\psi$ takes several gradient steps per encoder step.
Encoder, decoder, both priors, and $q_\psi$ are then trained jointly by stochastic gradient descent.
Stage~1 returns a disentangled variational autoencoder with an invariant latent kept fixed and a responsive block, which the stage~2 velocity field is trained to transport.

\subsection{Stage 2: Transporting the responsive block}
\label{sec:flow}

With the encoder frozen, stage~2 learns the conditional mechanism $p(\zn\mid\zi,u)$ in \eqref{eq:chain}. Given the invariant block $\zi$ of a control cell and a perturbation $u$, it generates the responsive block the cell would exhibit under $u$.
For each $u$, the responsive blocks of control cells and cells perturbed by $u$ define two distributions, and we learn a conditional flow that transports the former to the latter while keeping $\zi$ fixed.
This mirrors inference: the invariant block of an observed control cell is retained, and only its responsive block is generated.
The invariance penalty \eqref{eq:mi} supports this construction by encouraging $\zi$ to have the same distribution in control and perturbed cells.

\noindent \textbf{Conditional flow.}
Let $e_u$ be the perturbation code.
A velocity field $v_\theta(z,t\mid\zi,e_u)$ defines a transport through the ordinary differential equation $\dot z=v_\theta(z,t\mid\zi,e_u)$, integrated from $t{=}0$ to $t{=}1$.
We fit the field with conditional flow matching \citep{lipman2023flow,tong2024cfm}, a regression on the straight-line velocity between a source sample and a target sample that requires no simulation of the ordinary differential equation during training:
\begin{equation}
\mathcal{L}_{\mathrm{FM}}(\theta)=
\Exp_{t\sim\mathcal U[0,1],\,(z_0,z_1)\sim\pi}
\big\|\,v_\theta(z_t,t\mid \zi,e_u)-(z_1-z_0)\,\big\|^2,
\qquad z_t=(1-t)z_0+tz_1.
\label{eq:fm}
\end{equation}
The source $z_0=\zn^{\mathrm{ctrl}}$ is the responsive block of a control cell, the target $z_1=\zn^{\mathrm{pert}}$ is the responsive block of a cell under $u$, and $\zi$ is the invariant block of the control cell.
The field therefore displaces the responsive block of an observed cell.
The expectation in \eqref{eq:fm} is taken over pairs drawn from a coupling $\pi$ of the two populations.
Any such coupling yields a flow that carries the control distribution onto the perturbed distribution \citep{tong2024cfm}; the coupling determines which control cell is carried onto which perturbed cell.

\noindent \textbf{Pairing by minibatch entropic optimal transport.}
A perturbation experiment cannot record which perturbed cell originates from which control cell, so the pairs in \eqref{eq:fm} are unobserved.
We approximate such pairing from a minibatch entropic optimal transport plan, as in OT-CFM \citep{tong2024cfm}.
For each sampled perturbation $u$, we draw $n$ control and $n$ perturbed cells and solve
\begin{equation}
\pi^{\varepsilon}=\arg\min_{\pi\in\Pi_n}\;\sum_{i,j}\pi_{ij}\,C_{ij}-\varepsilon H(\pi),
\qquad
C_{ij}=\alpha_{\mathrm{nr}}\,\big\|\zi^{(i)}-\zi^{(j)}\big\|^2
+\alpha_{r}\,\big\|\zn^{(i)}-\zn^{(j)}\big\|^2,
\label{eq:ot}
\end{equation}
where $i$ indexes control cells, $j$ indexes perturbed cells, $\Pi_n$ is the set of couplings with uniform marginals, $H$ is the entropy, and $\varepsilon>0$ is the regularization strength.
We solve \eqref{eq:ot} with Sinkhorn iterations \citep{cuturi2013sinkhorn} and pair each perturbed cell with a control cell sampled from its column of $\pi^{\varepsilon}$.
The $\alpha_{\mathrm{nr}}$ term favors control cells with a similar invariant block, following the conditional optimal transport construction of \citet{kerrigan2024conditional}, so the difference within a pair is ascribed to the perturbation.
The $\alpha_{r}$ term favors pairs with a small displacement of the responsive block.
The plan is solved again on fresh cells at every minibatch round, and its only role is to supply the pairs that \eqref{eq:fm} regresses on. 
The finite minibatch and the entropic regularization make $\pi^{\varepsilon}$ an approximation of the optimal transport plan between the two populations. We stress that two distinct plans are at play: $\pi^{\varepsilon}$, approximated via minibatch entropic OT, only selects pairs for training the velocity field, whereas the velocity field itself learns the transport of the responsive block of control cells onto the responsive block of perturbed cells. At inference, this learned velocity field is used to transport the responsive block alone, via the scheme we detail next.

\subsection{Inference}
\label{sec:inference}

Predicting the response to a target perturbation $u^\star$ proceeds by drawing an observed control cell, encoding its invariant block $\zi$, integrating the flow conditioned on $(\zi,e_{u^\star})$ to generate a responsive block $\zn$, and decoding the pair.
The invariant state is copied from an observed control cell and only the response is generated.
The prediction is well posed provided the control invariant state lies in the support the field has seen under $u^\star$, which the invariance \eqref{eq:mi} encourages. Each stage-1 term shapes the responsive block for the stage-2 transport.
The conditional prior organizes $\zn$ by perturbation, which lets the flow interpolate to unseen $u$. The invariance term \eqref{eq:mi} concentrates the response in $\zn$ and makes the $\alpha_{\mathrm{nr}}$ term of the coupling a closer measure of cell identity.

\section{Experiments}
\label{sec:exp}

We evaluate \textsc{Drift} on perturbation benchmarks that test combinatorial and zero-shot
generalization to unseen perturbations. Norman \citep{norman2019} is a combinatorial benchmark with both single-gene and two-gene perturbations, and Replogle \citep{replogle2022mapping} is a zero-shot benchmark with single-gene perturbations. We compare \textsc{Drift} with multiple published methods from four paradigms. 1) Foundation models: scGPT \citep{cui2024scgpt}, scFoundation
\citep{hao2024large}, GeneCompass \citep{yang2024genecompass}, and CellFM
\citep{zeng2025cellfm}. 2) Graph-based models: GEARS \citep{roohani2024gears},
CellOracle \citep{kamimoto2023dissecting}, and GraphVCI \citep{wu2023graphvci}. 
3) Generative models: CPA \citep{lotfollahi2023cpa}, samsVAE \citep{bereket2023samsvae}, STATE
\citep{adduri2025predicting}, CellFlow \citep{klein2025cellflow}, scDFM
\citep{yu2026scdfm}, and scBIG
\citep{scbig2026}. 4) Statistical methods: Control, Linear, and Linear-scGPT
\citep{ahlmann2025deep}. 

We measure the \emph{shift} induced by a perturbation relative to control
\citep{ahlmann2025deep}. Following \citet{scbig2026}, we report nine metrics in three
groups: \emph{correlation} between predicted and true shifts over all genes
($\rho\Delta$) and the $20$ most differentially expressed genes
($\rho\Delta^{\mathrm{D}}$); \emph{directional agreement}, the fraction of genes with
correctly signed shifts ($\mathrm{ACC}\Delta$, $\mathrm{ACC}\Delta^{\mathrm{D}}$), together
with DES, the rank correlation between predicted and true log fold changes over the genes a
perturbation significantly moves; and \emph{discriminability}, measured by
PDS \citep{roohani2025virtualcell}.
PDS ranks the true condition among candidates by their distance to the prediction, thereby
penalizing responses that do not distinguish conditions despite high per-gene correlation.
We also report pointwise errors ($L_2$, MSE, and MAE). All results are averaged over five seeds; standard deviations are reported in Appendix~\ref{app:results-std}.
Definitions appear in
Appendix~\ref{app:metrics}, implementation details in
Appendix~\ref{app:impl}. 

\subsection{Results: Norman}

\textbf{Norman} \citep{norman2019} is a CRISPRa \citep{gilbert2014genome} screen in K562 cells,
where one or two genes are targeted and the transcriptome is measured in each cell. Norman is
a standard benchmark focusing on \emph{combinatorial} perturbation prediction:  it includes
both single-gene activations and gene pairs, allowing evaluation of whether a model can
predict a joint effect after observing its constituent genes separately. The \emph{additive} split holds out $32$ two-gene combinations whose constituent genes are each observed
individually during training, isolating combinatorial generalization without novel genes.
The \emph{holdout} split holds out $21$ single-gene and $5$ two-gene conditions.

\label{sec:res-additive}
\noindent \textbf{Additive.} Table~\ref{tab:norman-additive} compares \textsc{Drift} with the baselines on the additive
split. \textsc{Drift} achieves the best performance on all nine metrics and improves on the
strongest baseline, scBIG, globally on all metrics. The largest gains appear where
combinatorial generalization is more challenging. $\rho_\Delta^{D}$, which considers the
twenty genes most affected by each perturbation, improves by $8.3\%$; thus, the
gain is not driven by the many genes with negligible responses. PDS improves by
$7.2\%$ over scBIG, 
indicating that the predicted
shifts more clearly identify the condition that produced them. Pointwise errors
also improve substantially: MSE by $37.2\%$ and $L_2$ by $20.4\%$. Together,
these results show that \textsc{Drift} captures both the magnitude and the condition-specific
structure of combinatorial perturbation responses.

\begin{table*}[t]
\centering\footnotesize
\caption{Norman \textbf{additive} split. \textbf{Bold}: best; \underline{underline}: second best.}
\label{tab:norman-additive}
\setlength{\tabcolsep}{4pt}\renewcommand{\arraystretch}{1.1}
\resizebox{\textwidth}{!}{%
\begin{tabular}{|c|l||ccccccccc|}
\hline\thickhline
\rowcolor{mygray}
\textbf{Type} & \textbf{Method} & $\rho\Delta\uparrow$ & $\rho\Delta^{\mathrm{D}}\uparrow$ & $\mathrm{ACC}\Delta\uparrow$ & $\mathrm{ACC}\Delta^{\mathrm{D}}\uparrow$ & DES$\uparrow$ & PDS$\uparrow$ & L2$\downarrow$ & MSE$\downarrow$ & MAE$\downarrow$ \\
\hline\hline
\multirow{3}{*}{\rotatebox{90}{Sta.}} & Control & -- & -- & -- & -- & -- & 0.5000 & 9.25 & 0.0487 & 0.1720 \\
 & Linear & 0.6211 & 0.7075 & 0.8007 & 0.8906 & 0.3192 & 0.5363 & 6.76 & 0.0265 & 0.1233 \\
 & Linear-scGPT & 0.6925 & 0.7514 & 0.8424 & 0.9688 & 0.5023 & 0.7913 & 5.31 & 0.0155 & 0.0942 \\
\multirow{4}{*}{\rotatebox{90}{Found.}} & scGPT & 0.4408 & 0.4416 & 0.8125 & 0.4839 & 0.2504 & 0.5022 & 7.35 & 0.0296 & 0.1285 \\
 & scFoundation & 0.6813 & 0.4778 & 0.8768 & 0.7453 & 0.5409 & 0.7994 & 4.91 & 0.0138 & 0.0852 \\
 & GeneCompass & 0.6897 & 0.4810 & 0.8916 & 0.7484 & 0.5948 & 0.8024 & 4.77 & 0.0124 & 0.0808 \\
 & CellFM & 0.5947 & 0.5209 & 0.8819 & 0.8459 & 0.7550 & 0.7419 & 5.41 & 0.0161 & 0.0889 \\
\multirow{3}{*}{\rotatebox{90}{Gra.}} & GEARS & 0.7134 & 0.5065 & 0.8916 & 0.7641 & 0.6220 & 0.8155 & 4.59 & 0.0117 & 0.0788 \\
 & CellOracle & 0.0437 & 0.2567 & 0.4942 & 0.4536 & 0.0319 & 0.5277 & 11.10 & 0.0691 & 0.2001 \\
 & GraphVCI & 0.5468 & 0.5722 & 0.8034 & 0.7905 & 0.6495 & 0.5151 & 6.68 & 0.0258 & 0.1177 \\
\multirow{5}{*}{\rotatebox{90}{Gen.}} & samsVAE & 0.6497 & 0.7750 & 0.8470 & 0.9658 & 0.6138 & 0.6689 & 6.87 & 0.0329 & 0.1067 \\
 & STATE & 0.2439 & 0.2628 & 0.8023 & 0.4141 & 0.4891 & 0.5171 & 9.02 & 0.0421 & 0.1343 \\
 & CellFlow & 0.7892 & 0.7275 & 0.9151 & 0.9391 & 0.8113 & 0.7581 & 4.54 & 0.0114 & 0.0775 \\
 & scBIG & \underline{0.8496} & \underline{0.8230} & \underline{0.9197} & \underline{0.9906} & \underline{0.8593} & \underline{0.8548} & \underline{3.92} & \underline{0.0091} & \underline{0.0689} \\
\rowcolor{mygray1}
 & \textbf{DRIFT (ours)} & \textbf{0.8871} & \textbf{0.8914} & \textbf{0.9222} & \textbf{0.9977} & \textbf{0.9017} & \textbf{0.9167} & \textbf{3.12} & \textbf{0.0057} & \textbf{0.0549} \\
\hline
\end{tabular}}
\end{table*}

\noindent \textbf{Holdout split.} \label{sec:res-holdout} The holdout split is even more challenging: nine of its $26$ test conditions involve
a gene absent from training, requiring zero-shot generalization;
the double perturbations additionally test combinatorial generalization.
Table~\ref{tab:norman-holdout} reports results separately for single and double
perturbations. On \emph{singles}, \textsc{Drift} outperforms scBIG on all nine metrics:
$\rho_\Delta^{\mathrm{D}}$ rises by $7.8\%$, 
while
MSE improves by $4.6\%$.  
These gains hold despite the
need to extrapolate to gene perturbations unseen during training.
On \emph{doubles}, the additive-split pattern reappears more strongly: \textsc{Drift}       
outperforms scBIG on every metric by a wider margin than on singles.
$\rho_\Delta^{\mathrm{D}}$ improves $6.9\%$ 
and
$\mathrm{ACC}\Delta^{\mathrm{D}}$ nearly saturates at $0.9873$. PDS improves $8.5\%$,
and the pointwise errors improve furthest of all,
MSE by $28.9\%$ and $L_2$ by $15.8\%$. These results show that 
\textsc{Drift} superiority carries over when the constituent genes are observed,
even within the more challenging holdout regime.

\begin{table*}[t]
\centering\footnotesize
\caption{Norman \textbf{holdout} split. \textbf{Bold}: best; \underline{underline}: second best.}
\label{tab:norman-holdout}
\setlength{\tabcolsep}{4pt}\renewcommand{\arraystretch}{1.1}
\resizebox{\textwidth}{!}{%
\begin{tabular}{|c|l||ccccccccc|}
\hline\thickhline
\rowcolor{mygray}
\textbf{Type} & \textbf{Method} & $\rho\Delta\uparrow$ & $\rho\Delta^{\mathrm{D}}\uparrow$ & $\mathrm{ACC}\Delta\uparrow$ & $\mathrm{ACC}\Delta^{\mathrm{D}}\uparrow$ & DES$\uparrow$ & PDS$\uparrow$ & L2$\downarrow$ & MSE$\downarrow$ & MAE$\downarrow$ \\
\hline\hline
\multicolumn{11}{|c|}{\textit{Single gene perturbation}} \\
\hline\hline
\multirow{3}{*}{\rotatebox{90}{Sta.}} & Control & -- & -- & -- & -- & -- & 0.5000 & 4.33 & 0.0108 & 0.0685 \\
 & Linear & 0.5924 & 0.6021 & 0.7695 & 0.8476 & 0.4548 & 0.5119 & 3.46 & 0.0067 & 0.0516 \\
 & Linear-scGPT & 0.6072 & 0.6352 & 0.7620 & 0.8095 & 0.4584 & 0.6571 & 3.60 & 0.0072 & 0.0575 \\
\multirow{4}{*}{\rotatebox{90}{Found.}} & scGPT & 0.5172 & 0.5358 & 0.7509 & 0.8286 & 0.5987 & 0.5048 & 3.58 & 0.0071 & 0.0530 \\
 & scFoundation & 0.4453 & 0.3144 & 0.6882 & 0.7429 & 0.3897 & 0.6619 & 3.65 & 0.0079 & 0.0568 \\
 & GeneCompass & 0.5307 & 0.3423 & 0.7006 & 0.7429 & 0.4543 & 0.5857 & 3.44 & 0.0066 & 0.0546 \\
 & CellFM & 0.4331 & 0.5226 & 0.6708 & 0.7119 & 0.5318 & 0.5524 & 3.83 & 0.0086 & 0.0600 \\
\multirow{2}{*}{\rotatebox{90}{Gra.}} & GEARS & 0.4539 & 0.5345 & 0.7059 & 0.8024 & 0.3968 & 0.4905 & 3.77 & 0.0078 & 0.0574 \\
 & GraphVCI & 0.3902 & 0.1865 & 0.7054 & 0.5583 & 0.1395 & 0.5000 & 3.28 & 0.0066 & 0.0500 \\
\multirow{5}{*}{\rotatebox{90}{Gen.}} & samsVAE & 0.4778 & 0.4982 & 0.7050 & 0.8310 & 0.3366 & 0.3452 & 3.88 & 0.0081 & 0.0614 \\
 & STATE & 0.3085 & 0.5207 & 0.7600 & 0.5763 & 0.5712 & 0.5048 & 5.94 & 0.0177 & 0.0737 \\
 & CellFlow & 0.5930 & 0.4736 & 0.7778 & 0.8119 & 0.6280 & 0.6643 & 3.40 & 0.0076 & 0.0535 \\
 & scBIG & \underline{0.6507} & \underline{0.6537} & \underline{0.7876} & \underline{0.8548} & \underline{0.6333} & \underline{0.7467} & \underline{3.26} & \underline{0.0065} & \underline{0.0491} \\
\rowcolor{mygray1}
 & \textbf{DRIFT (ours)} & \textbf{0.6690} & \textbf{0.7046} & \textbf{0.7890} & \textbf{0.8783} & \textbf{0.6904} & \textbf{0.7596} & \textbf{3.09} & \textbf{0.0062} & \textbf{0.0480} \\
\hline\hline
\multicolumn{11}{|c|}{\textit{Double gene perturbations}} \\
\hline\hline
\multirow{3}{*}{\rotatebox{90}{Sta.}} & Control & -- & -- & -- & -- & -- & 0.5000 & 6.02 & 0.0200 & 0.0984 \\
 & Linear & 0.7179 & 0.8277 & 0.8240 & 0.9200 & 0.5637 & 0.5000 & 4.31 & 0.0105 & 0.0656 \\
 & Linear-scGPT & 0.7351 & 0.7230 & 0.8282 & 0.9100 & 0.6118 & 0.7000 & 3.91 & \underline{0.0080} & 0.0626 \\
\multirow{4}{*}{\rotatebox{90}{Found.}} & scGPT & 0.6350 & 0.4719 & 0.8068 & 0.8333 & 0.4456 & 0.5000 & 4.70 & 0.0124 & 0.0726 \\
 & scFoundation & 0.6605 & 0.4445 & 0.7964 & 0.9200 & 0.6538 & 0.6500 & 3.91 & 0.0083 & 0.0589 \\
 & GeneCompass & 0.6437 & 0.6866 & 0.7471 & 0.7400 & 0.5905 & 0.6000 & 4.37 & 0.0108 & 0.0723 \\
 & CellFM & 0.5615 & 0.8151 & 0.7072 & 0.7600 & 0.6746 & 0.5000 & 4.89 & 0.0132 & 0.0804 \\
\multirow{2}{*}{\rotatebox{90}{Gra.}} & GEARS & 0.5087 & 0.6578 & 0.7130 & 0.8300 & 0.4156 & 0.5000 & 5.00 & 0.0138 & 0.0788 \\
 & GraphVCI & 0.4290 & 0.5429 & 0.6889 & 0.6533 & 0.4003 & 0.5048 & 4.61 & 0.0114 & 0.0725 \\
\multirow{5}{*}{\rotatebox{90}{Gen.}} & samsVAE & 0.6267 & 0.6402 & 0.7539 & 0.8800 & 0.5183 & 0.6500 & 4.66 & 0.0118 & 0.0742 \\
 & STATE & 0.3786 & 0.6379 & 0.8126 & 0.5800 & 0.7097 & 0.5000 & 5.94 & 0.0177 & 0.0737 \\
 & CellFlow & 0.7404 & 0.6774 & 0.8388 & 0.8400 & 0.7636 & 0.4500 & 4.12 & 0.0097 & 0.0621 \\
 & scBIG & \underline{0.8026} & \underline{0.8719} & \underline{0.8552} & \underline{0.9700} & \underline{0.8313} & \underline{0.7440} & \underline{3.68} & 0.0083 & \underline{0.0554} \\
\rowcolor{mygray1}
 & \textbf{DRIFT (ours)} & \textbf{0.8562} & \textbf{0.9319} & \textbf{0.8724} & \textbf{0.9873} & \textbf{0.8734} & \textbf{0.8075} & \textbf{3.10} & \textbf{0.0059} & \textbf{0.0469} \\
\hline
\end{tabular}}
\end{table*}

\subsection{Results: Replogle RPE1}
\label{sec:res-replogle}

\textbf{Replogle2022-RPE1} \citep{replogle2022mapping} is a CRISPRi knockdown screen in RPE1 cells. Compared with Norman, it differs in cell type, perturbation modality (knockdown rather than activation), and scale ($1{,}037$ training conditions versus $190$). All $384$ test conditions are single-gene perturbations targeting genes absent from training, making this a fully \emph{zero-shot} setting. We use the split of \citet{scbig2026}. DRIFT leads on all nine metrics (Table~\ref{tab:norman-replogle}).
 It surpasses scBIG on both correlation metrics $\rho_\Delta$ and $\rho_\Delta^{\mathrm{D}}$ and PDS. Its largest gain is in
DES ($+36.5\%$),  
the rank agreement between predicted and true log fold
changes over the genes a perturbation significantly moves. These results provide strong evidence that the
disentangled formulation captures generalizable structure in perturbation responses across several datasets.

\begin{table*}[t]
\centering\footnotesize
\caption{Replogle RPE1. \textbf{Bold}: best; \underline{underline}: second best.}
\label{tab:norman-replogle}
\setlength{\tabcolsep}{4pt}\renewcommand{\arraystretch}{1.1}
\resizebox{\textwidth}{!}{%
\begin{tabular}{|c|l||ccccccccc|}
\hline\thickhline
\rowcolor{mygray}
\textbf{Type} & \textbf{Method} & $\rho\Delta\uparrow$ & $\rho\Delta^{\mathrm{D}}\uparrow$ & $\mathrm{ACC}\Delta\uparrow$ & $\mathrm{ACC}\Delta^{\mathrm{D}}\uparrow$ & DES$\uparrow$ & PDS$\uparrow$ & L2$\downarrow$ & MSE$\downarrow$ & MAE$\downarrow$ \\
\hline\hline
\multirow{3}{*}{\rotatebox{90}{Sta.}} & Control & -- & -- & -- & -- & -- & 0.5000 & 7.03 & 0.0171 & 0.0768 \\
 & Linear & 0.2720 & 0.3630 & 0.5816 & 0.6919 & 0.2149 & 0.5318 & 14.13 & 0.0916 & 0.1491 \\
 & Linear-scGPT & 0.3404 & 0.5438 & 0.5852 & 0.7681 & 0.2497 & 0.5338 & 7.60 & 0.0172 & 0.0890 \\
\multirow{4}{*}{\rotatebox{90}{Found.}} & scGPT & 0.2840 & 0.5691 & 0.5882 & 0.7759 & 0.2980 & 0.5001 & 9.49 & 0.0246 & 0.1037 \\
 & scFoundation & 0.2037 & 0.3632 & 0.5489 & 0.6908 & 0.2331 & 0.5098 & 7.07 & 0.0155 & 0.0812 \\
 & GeneCompass & 0.3113 & 0.4299 & 0.5758 & 0.7148 & 0.2873 & 0.5038 & 7.18 & 0.0162 & 0.0822 \\
 & CellFM & 0.3060 & 0.4299 & 0.5733 & 0.7086 & 0.2819 & 0.5039 & 7.20 & 0.0163 & 0.0825 \\
\multirow{2}{*}{\rotatebox{90}{Gra.}} & GEARS & 0.3083 & 0.4289 & 0.5752 & 0.7096 & 0.2820 & 0.5004 & 7.18 & 0.0162 & 0.0822 \\
 & GraphVCI & 0.4294 & 0.5302 & 0.5589 & 0.6436 & 0.1863 & 0.5000 & 7.41 & 0.0157 & 0.0913 \\
\multirow{5}{*}{\rotatebox{90}{Gen.}} & samsVAE & 0.4083 & 0.5204 & 0.5998 & 0.7565 & 0.3280 & 0.4984 & 6.86 & 0.0140 & 0.0780 \\
 & STATE & 0.3638 & 0.5298 & 0.6463 & 0.7891 & 0.3711 & 0.5011 & 8.38 & 0.0194 & 0.0759 \\
 & CellFlow & 0.3878 & 0.4391 & 0.6226 & 0.7194 & 0.3114 & 0.5441 & 6.48 & 0.0135 & 0.0711 \\
 & scBIG & \underline{0.4875} & \underline{0.5925} & \underline{0.6471} & \underline{0.8089} & \underline{0.5145} & \underline{0.5520} & \underline{6.12} & \underline{0.0118} & \underline{0.0676} \\
\rowcolor{mygray1}
 & \textbf{DRIFT (ours)} & \textbf{0.5085} & \textbf{0.6160} & \textbf{0.6537} & \textbf{0.8149} & \textbf{0.7021} & \textbf{0.6044} & \textbf{5.68} & \textbf{0.0101} & \textbf{0.0633} \\
\hline
\end{tabular}}
\end{table*}

\subsection{Results: ComboSciPlex}
\label{sec:res-combo}

\textbf{ComboSciPlex} \citep{srivatsan2020sciplex} is a drug molecule perturbation benchmark with 
$63{,}378$ cells exposed to $17$ compounds. Of its $31$ perturbed conditions, $25$ are compound
pairs; $24$ conditions are used for training and $7$ are held out. We follow the
protocol of \citet{yu2026scdfm} and report the two metrics it uses in place of DES and PDS: DE-Spearman $\rho$, the rank correlation between predicted and measured log fold changes on the statistically significant differentially expressed genes, and the discrimination score (DS) of \citet{roohani2025virtualcell}, which measures whether the predicted populations of different perturbations remain separated from one another.
\textsc{Drift} leads on five of the six metrics. Against scDFM, the strongest baseline, it reduces
pointwise errors by $23$ to $33\%$ and raises DE-Spearman $\rho$ from $0.829$ to $0.890$
and $\rho\Delta$ from $0.893$ to $0.947$. In DS, DRIFT ranks ahead of scDFM and slightly below CPA, which performs worse on all remaining
metrics. These results suggest that our formulation extends beyond genetic perturbations to combinations of drugs.

\begin{table*}[t]
\centering\footnotesize
\caption{ComboSciPlex. \textbf{Bold}: best; \underline{underline}: second best.}
\label{tab:combo-scdfm}
\setlength{\tabcolsep}{6pt}\renewcommand{\arraystretch}{1.1}
\begin{tabular}{|l||cccccc|}
\hline\thickhline
\rowcolor{mygray}
\textbf{Method} & $\rho\Delta\uparrow$ & DE-Spearman $\rho\uparrow$ & DS$\uparrow$ &
L2$\downarrow$ & MSE$\downarrow$ & MAE$\downarrow$ \\
\hline\hline
Control & N.A. & N.A. & 0.5714 & 5.3716 & 0.0324 & 0.0698 \\
scGPT & 0.8322 & $-0.1261$ & 0.8571 & 1.6934 & 0.0031 & 0.0251 \\
CPA & 0.8150 & 0.7906 & \textbf{0.8980} & 1.6592 & 0.0029 & 0.0240 \\
scDFM & \underline{0.8933} & \underline{0.8289} & 0.8776 & \underline{1.6567} & \underline{0.0028} & \underline{0.0220} \\
\hline
\rowcolor{mygray1}
\textbf{DRIFT (ours)} & \textbf{0.9467} & \textbf{0.8901} & \underline{0.8816} & \textbf{1.2804} & \textbf{0.0019} & \textbf{0.0150} \\
\hline
\end{tabular}
\end{table*}

\subsection{Ablations} 
We assess the contribution of each \textsc{Drift} component through ablation studies (Appendix~\ref{app:ablations}).
Disentanglement drives most of the performance gains, improving every metric across datasets
and making the largest contribution overall. The invariance penalty
provides consistent complementary gains. Conditioning regularization is especially valuable in
zero-shot settings, where test perturbations are entirely unseen, but is less consequential for
combinatorial generalization, where test perturbations combine genes observed during training (details in Table~\ref{tab:ablation-components}).

\section{Conclusion}
\label{sec:conclusion}

We presented \textsc{Drift}, a two-stage model that disentangles a cell's invariant state from its
perturbation-responsive component, then transports the latter while preserving the former. In the
first stage, a variational autoencoder learns this disentangled representation through conditional
priors informed by distinct auxiliary variables and an information-theoretic invariance constraint,
thereby partitioning the latent state into invariant and responsive blocks. In the second stage, a conditional flow transports only the responsive component by learning from data a
velocity field that captures perturbation-induced dynamics in the
responsive latent block.

Across the Norman additive and holdout splits, the Replogle RPE1 screen, and the ComboSciPlex drug perturbation dataset, \textsc{Drift}
outperforms the strongest published baseline. These improvements are consistent across tasks, on both genetic and chemical perturbation, from learning combinatorial effects from single gene perturbations to generalizing to unseen perturbations, and across correlation, directional agreement, and discriminability metrics. Together, these results suggest that the disentangled formulation captures a central feature of perturbation response: separating invariant and responsive components reduces confounding from cell-level variation and yields a more faithful representation of perturbation-induced dynamics.

\paragraph{Limitations.} Although \textsc{Drift} is designed to accommodate substantial cell-to-cell variation,
current benchmarks widely adopted in the literature are limited to single cell lines. Extending the model to multiple cell lines would test whether the invariant--responsive
decomposition generalizes across diverse cellular backgrounds. Temporal measurements could also provide a
biological interpretation for the flow's time coordinate, capture response dynamics, and connect endpoint
mappings into trajectory models.

\newpage
\subsection*{AI USE STATEMENT}
In this work, we used generative AI tools to assist with \LaTeX{} table formatting, TikZ figure refinement, and grammar and style editing; to generate figures and tables from stored results; to support code design, refinement, and unit testing; to facilitate low-level interaction with the GPU cluster scheduler; and to manage artifacts, checkpoints, and datasets. We take full responsibility for the final content of this work, including all text, claims, and artifacts produced with the assistance of generative AI.

\subsection*{REPRODUCIBILITY STATEMENT}

All datasets required to reproduce the results in this paper are publicly available through the official repositories of the referenced works \cite{scbig2026,yu2026scdfm}.
Code and scripts for reproducing all reported results, including the training and evaluation of \textsc{DRIFT}, are available at \url{https://github.com/MustaphaBounoua/drift}.

\bibliographystyle{iclr2027_conference}
\bibliography{references}
\newpage
\appendix
\section{Metric definitions}
\label{app:metrics}

We follow the definitions and released implementation of \citet{scbig2026}. For a perturbation
$p\in\mathcal P$, $\mathbf x_p,\hat{\mathbf x}_p\in\Real^{G}$ are the observed and predicted mean
expression profiles, $\bar{\mathbf x}_{\mathrm{ctrl}}$ is the mean of the training control cells,
and $\mathcal D_p$ is the set of the $20$ genes with the largest
$|x_{pg}-\bar x_{\mathrm{ctrl},g}|$.

\paragraph{Point-prediction accuracy.}
\begin{equation}
\mathrm{MSE}=\frac{1}{|\mathcal P|G}\sum_{p}\|\mathbf x_p-\hat{\mathbf x}_p\|_2^2,\qquad
\mathrm{MAE}=\frac{1}{|\mathcal P|G}\sum_{p}\|\mathbf x_p-\hat{\mathbf x}_p\|_1,\qquad
L_2=\frac{1}{|\mathcal P|}\sum_{p}\|\mathbf x_p-\hat{\mathbf x}_p\|_2 .
\end{equation}

\paragraph{Biological response fidelity.}
\begin{equation}
\rho\Delta=\frac{1}{|\mathcal P|}\sum_{p}\operatorname{Corr}\big(\hat{\mathbf x}_p-\bar{\mathbf x}_{\mathrm{ctrl}},\ \mathbf x_p-\bar{\mathbf x}_{\mathrm{ctrl}}\big),
\end{equation}
\begin{equation}
\mathrm{ACC}\Delta=\frac{1}{|\mathcal P|G}\sum_{p}\sum_{g=1}^{G}\mathbb I\big[\operatorname{sign}(\hat x_{pg}-\bar x_{\mathrm{ctrl},g})=\operatorname{sign}(x_{pg}-\bar x_{\mathrm{ctrl},g})\big],
\end{equation}
where $\operatorname{Corr}$ is the Pearson correlation; $\rho\Delta^{\mathrm D}$ and
$\mathrm{ACC}\Delta^{\mathrm D}$ restrict both to $\mathcal D_p$. DES is the Spearman correlation of
log fold changes over the genes identified as significant by a Wilcoxon rank-sum test between
perturbed and control cells ($p<0.05$):
\begin{equation}
\mathrm{DES}=\frac{1}{|\mathcal P|}\sum_{p}\operatorname{Spearman}\Big(\log_2\frac{\hat{\mathbf x}_{p,\mathrm{sig}}}{\bar{\mathbf x}_{\mathrm{ctrl},\mathrm{sig}}},\ \log_2\frac{\mathbf x_{p,\mathrm{sig}}}{\bar{\mathbf x}_{\mathrm{ctrl},\mathrm{sig}}}\Big).
\end{equation}

\paragraph{Discriminative power.}
PDS ranks the observed profiles of all perturbations in $\mathcal P$ by their $L_1$ distance to
$\hat{\mathbf x}_p$, excluding the genes targeted by $p$; with $\mathrm{rank}_p$ the position of
$\mathbf x_p$,
\begin{equation}
\mathrm{PDS}=\frac{1}{|\mathcal P|}\sum_{p}\Big(1-\frac{\mathrm{rank}_p-1}{|\mathcal P|-1}\Big).
\end{equation}

\paragraph{Distributional metrics.}
Following \cite{scbig2026}, we assess distributional fidelity using three metrics that compare the predicted and observed cell \emph{populations} for each perturbation, rather
than their means, on cells projected onto $50$ principal components. \textbf{E-Dist} is the energy
distance between the two populations: zero exactly when they coincide, and sensitive to their
spread as well as their location. \textbf{Wasserstein} is the entropic optimal transport cost
($\varepsilon=1$, uniform weights) between them, the average distance cells must travel to turn one
population into the other. \textbf{Discr.~Cos} is the cosine between the true and predicted mean
shift from control, and so measures only the direction of the response, not its magnitude. All
three are averaged over $\mathcal P$.

\paragraph{ComboSciPlex.}
We evaluate ComboSciPlex using \texttt{cell-eval}, following \citet{yu2026scdfm}.

\section{Implementation}
\label{app:impl}

\paragraph{Datasets.} Norman \citep{norman2019} uses CRISPRa to activate $105$ genes and $131$
gene pairs in K562 cells; Replogle RPE1 \citep{replogle2022mapping} uses CRISPRi to silence
essential genes. We use the preprocessing and splits of \citet{scbig2026}:
$\ln(\mathrm{CPM}{+}1)$ on highly variable genes ($2051$ for Norman; $3754$ for Replogle),
and follow the same train/validation/test splits as \citet{scbig2026}, ensuring directly comparable
evaluation. Baseline results on Norman and Replogle are reported from
\citet{scbig2026}, on ComboSciPlex they are reported from \citet{yu2026scdfm}. Norman's \emph{additive} split trains on $167$ conditions
and holds out $32$ pairs whose constituent genes are observed singly, testing combinatorial
generalization. Its \emph{holdout} split trains on $190$ conditions and holds out $26$ that include genes
absent from training, alone and in pairs, testing zero-shot generalization. Replogle holds out $384$ unseen-gene
single-perturbation conditions. ComboSciPlex \citep{srivatsan2020sciplex} profiles
A549 cells under $17$ compounds across $31$ conditions ($24$ train, $7$ test); following
\citet{yu2026scdfm}, we use the same train/validation/test splits, train on $5000$ highly variable genes, evaluate on $1000$ test-selected
genes, and generate $128$ cells per condition. 

\paragraph{Stage 1: the disentangling VAE.} The encoder and decoder are MLPs on
$\ln(\mathrm{CPM}{+}1)$ expression. The encoder receives the perturbation $u$ and the covariates $c$
and emits the invariant block $z_{\mathrm{nr}}$ and the responsive block $z_r$; $u$ also conditions
the responsive prior and $c$ the invariant prior. The decoder has a single mean-squared-error head,
used for training and for every evaluation. $c$ comprises the sequencing run together with per-cell sequencing depth, genes detected, and mitochondrial and ribosomal content; we verify that these are close to perturbation-independent. On ComboSciPlex $c$ is empty: the screen is arrayed, so every well receives a single drug condition and the well label is not independent of $u$. Table~\ref{tab:hyperparams} gives all settings.

The perturbation encoder is additive with an interaction term:
$s_0=\varphi(g_1)+\varphi(g_2)$, $s=s_0+\psi([\,s_0,\ \varphi(g_1)\odot\varphi(g_2)\,])$
and $e_u=\rho(s)$, with $\varphi$ an MLP. Both
the sum and the elementwise product are symmetric in $(g_1,g_2)$, so the code does not
depend on gene order. The interaction branch $\psi$ is applied only when both genes are
present, and is zero-initialized, so composition begins exactly additive and becomes
interactive only if the data pay for it; a control cell returns a learned \textsc{null}
code and a gene with no feature row is routed to a learned \textsc{unknown}.

\paragraph{Perturbation features.} For a genetic perturbation, each gene is represented by the concatenation of
three blocks: ESM2 protein embeddings \citep{lin2023esm2} reduced by PCA
fitted over the full gene dictionary; a neighborhood in the STRING network of protein--protein
associations \citep{szklarczyk2023string}, taken over
the whole interactome; and co-membership in Reactome biological pathways \citep{milacic2024reactome}.  We use these features following the official implementation of \citet{scbig2026}. The two
network blocks describe a gene by what it interacts with and what processes it belongs to, which
the protein sequence alone does not carry. Genes absent from a source receive an exactly zero block
in that source. Table~\ref{tab:ablation-features} measures what the two network blocks
contribute over the protein sequence alone. For the chemical perturbation, each compound is represented by a Morgan fingerprint \citep{rogers2010ecfp} of
radius $4$ and $1024$ bits, the representation and settings used for molecule
perturbations by \citet{klein2025cellflow}, reduced by PCA to $16$
dimensions.

\paragraph{Objective.} Both KL terms are warmed up linearly, and the CLUB penalty on
$I(z_{\mathrm{nr}};u)$ follows the same schedule, so it does not constrain an untrained latent. The
CLUB critic predicts a projection of the condition code $e_u$ from $z_{\mathrm{nr}}$ and is trained
with its own optimizer at a learning rate of $10^{-3}$.

\paragraph{Stage 2: the conditional flow.} The velocity field is a DiT in which the two perturbed
genes enter as separate tokens, and the stage-1 perturbation encoder is inherited. Training pairs
come from an entropic optimal transport plan (Sinkhorn) over blocks of cells drawn from several
conditions, with the cost of \eqref{eq:ot}. Generation integrates the probability-flow ODE with
Euler steps, using the EMA weights.

\begin{table}[t]
\centering\footnotesize
\caption{Configuration of DRIFT, read from the resolved configuration of the checkpoints that
produced the tables. A single centered entry means the setting is shared by the three datasets.}
\label{tab:hyperparams}
\setlength{\tabcolsep}{5pt}\renewcommand{\arraystretch}{1.05}
\newcommand{\allthree}[1]{\multicolumn{3}{c}{#1}}
\begin{tabular}{llccc}
\hline\thickhline
\rowcolor{mygray}
\textbf{Category} & \textbf{Hyperparameter} & \textbf{Norman} & \textbf{Replogle} & \textbf{ComboSciPlex} \\
\hline\hline
\multirow{9}{*}{\rotatebox{90}{Model}}
& Invariant block $\di$ & \allthree{64} \\
& Responsive block $\dn$ & \allthree{192} \\
& Encoder and decoder width & \allthree{1024 ($2$ layers)} \\
& Condition code $d_{\mathrm{pert}}$ & \allthree{128} \\
& Perturbation encoder width & \allthree{256} \\
& Perturbation features & $3\times128$ & $3\times128$ & 16 \\
& Velocity field (DiT) blocks / tokens & \allthree{6 / 16} \\
& Conditioning width & \allthree{256} \\
& Time embedding frequencies & \allthree{128} \\
\hline
\multirow{6}{*}{\rotatebox{90}{Stage 1}}
& Epochs & \allthree{120} \\
& Batch size & \allthree{256} \\
& Optimizer / learning rate & \allthree{Adam / $10^{-4}$} \\
& $\beta_r$, $\beta_{\mathrm{nr}}$ & \allthree{0.5, 4} \\
& KL and MI warm-up & \allthree{20 epochs} \\
& $\lambda_{\mathrm{CLUB}}$ / critic dimension / inner steps & \allthree{5 / 32 / 5} \\
\hline 
\multirow{8}{*}{\rotatebox{90}{Stage 2}}
& Steps & \allthree{60{,}000} \\
& Batch size & \allthree{512} \\
& Optimizer / learning rate & \allthree{Adam / $10^{-4}$} \\
& EMA decay & \allthree{0.999} \\
& OT block: cells $\times$ conditions & \allthree{$512\times16$} \\
& Sinkhorn $\varepsilon$; $\alpha_{\mathrm{nr}}$, $\alpha_r$ & \allthree{0.5; 1, 1} \\
& Euler steps at inference & \allthree{50} \\
\hline
\end{tabular}
\end{table}

\begin{algo}{Inference: predicting the response to a perturbation $u^\star$}
\label{alg:infer}
\item \textbf{Input:} target perturbation $u^\star$; observed control cells; number of cells $N$.
\item \textbf{Output:} predicted perturbed cells $\{\hat x_i\}_{i=1}^{N}$.
\item $e_{u^\star}\leftarrow$ perturbation encoder applied to $u^\star$
\item \textbf{for} $i=1$ \textbf{to} $N$
\item \algin draw a control cell $x$ and encode it, $(\zi,\zn)\leftarrow q_\phi(x,u_{\mathrm{ctrl}},c)$
\item \algin integrate $\dot z=v_\theta(z,t\mid\zi,e_{u^\star})$ from $z_0=\zn$ to $t=1$
      ($50$ Euler steps), giving $\zn'=z_1$
\item \algin $\hat x_i\leftarrow p_\theta([\zi,\zn'])$, with $\zi$ copied from the control cell
\item \textbf{end for}
\end{algo}

\section{Conditioning regularization}
\label{app:condpath}

\paragraph{Response supervision.} A linear head $h:\mathbb R^{d_{\mathrm{pert}}}
\!\to\mathbb R^{G}$ predicts the condition's mean training response
$\bar\Delta_c=\bar x_c-\bar x_{\mathrm{ctrl}}$,
\begin{equation}
\mathcal L_{\mathrm{aux}}
=\frac{1}{|\mathcal C_{\mathrm{tr}}|}\sum_{c\in\mathcal C_{\mathrm{tr}}}
\big\lVert h(e_u^{(c)})-\bar\Delta_c\big\rVert_2^2 ,
\label{eq:aux}
\end{equation}
which ties $e_u$ to the cell responses it must eventually explain. Without it $\varphi$
is shaped only by what reaches it through the KL term of the responsive prior, a weak
and indirect signal. The head is kept linear so that it cannot itself absorb the
structure we are inducing in $e_u$, and is not used after training.

\paragraph{Relative isometry.} Let $D^{e}_{ij}=\lVert e_u^{(i)}-e_u^{(j)}\rVert$ and
$D^{\Delta}_{ij}=\lVert\bar\Delta_i-\bar\Delta_j\rVert$ be the pairwise distances between codes and
between mean training responses. We penalize the departure of their correlation from one,
\begin{equation}
\mathcal L_{\mathrm{iso}}=1-\mathrm{corr}\big(\operatorname{vec}D^{e},\ \operatorname{vec}D^{\Delta}\big).
\label{eq:iso}
\end{equation}
This gives perturbations with similar responses similar codes.
Matching relative distances instead of absolute ones constrains only the ordering and
spacing of conditions, leaving the scale of $e_u$ free to adapt to the prior.

\paragraph{Noise augmentation.} During training, Gaussian noise is added to the condition embedding.
This prevents the representation from collapsing into a lookup table over training conditions and
ensures that neighborhoods around learned codes remain valid, so unseen genes near seen
ones can still be decoded sensibly.

\section{What each latent block encodes}
\label{app:viz}

\begin{figure}[t]
\centering
\includegraphics[width=\textwidth]{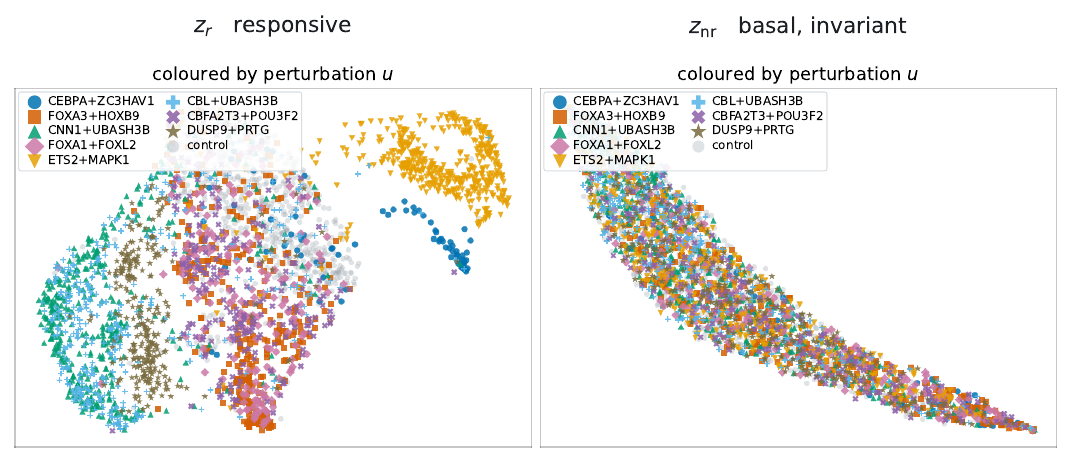}
\caption{Norman additive: UMAPs of the responsive block $z_r$ and invariant block
$z_{\mathrm{nr}}$, colored by perturbation.}
\label{fig:latent-additive}
\end{figure}

\begin{figure}[t]
\centering
\includegraphics[width=\textwidth]{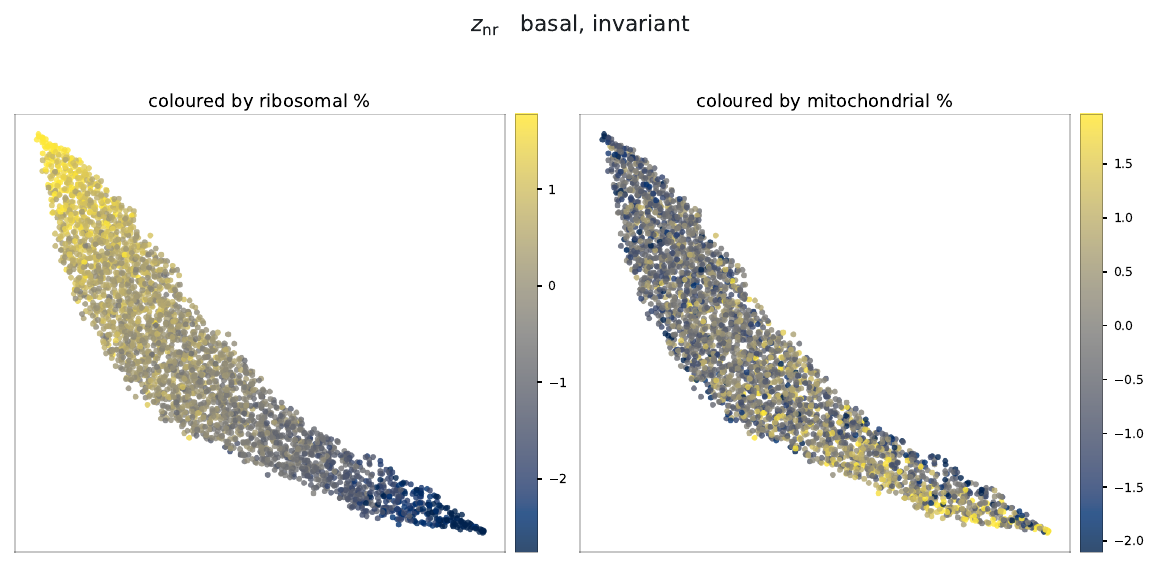}
\caption{The invariant block of Figure~\ref{fig:latent-additive}, colored by the numeric covariates
in $c$.}
\label{fig:latent-covar}
\end{figure}

Figure~\ref{fig:latent-additive} shows that the responsive block groups cells by perturbation,
whereas the invariant block interleaves them and captures covariates in $c$
(Figure~\ref{fig:latent-covar}). Table~\ref{tab:latent-probe} confirms this separation across
vocabularies, from the $20$ strongest perturbations to all conditions: $z_r$ predicts perturbation
identity accurately, while $z_{\mathrm{nr}}$ remains far below $z_r$. The same pattern
holds for mutual information: $z_r$ captures most of $u$, whereas $I(z_{\mathrm{nr}},u)$ is very low. This ordering is consistent across linear and MLP probes, as well as variational and MINE
estimators.

\begin{table*}[t]
\centering\footnotesize
\caption{Decoding \emph{which} perturbation a cell received from each latent block. Classes are
balanced, so chance is $1/K$. MINE \cite{belghazi2018mutual} is a mutual-information lower bound
in nats.}
\label{tab:latent-probe}
\setlength{\tabcolsep}{5pt}\renewcommand{\arraystretch}{1.1}
\begin{tabular}{|l|l|c||cc|cc|cc|}
\hline\thickhline
\rowcolor{mygray}
 & & & \multicolumn{4}{c|}{\textbf{probe accuracy}} & \multicolumn{2}{c|}{\textbf{MINE (nats)}} \\
\rowcolor{mygray}
 & & & \multicolumn{2}{c|}{linear} & \multicolumn{2}{c|}{MLP} & & \\
\rowcolor{mygray}
\textbf{Dataset} & \textbf{Conditions} & \textbf{chance} &
$z_{\mathrm{nr}}$ & $z_r$ & $z_{\mathrm{nr}}$ & $z_r$ & $z_{\mathrm{nr}}$ & $z_r$ \\
\hline\hline
 \multirow{4}{*}{Norman additive} & top 20 & 0.0500 & 0.157 & \textbf{0.981} & 0.171 & \textbf{0.991} & 0.14 & \textbf{3.35} \\
  & top 50 & 0.0200 & 0.100 & \textbf{0.994} & 0.093 & \textbf{0.998} & 0.15 & \textbf{4.00} \\
  & top 100 & 0.0100 & 0.053 & \textbf{0.980} & 0.056 & \textbf{0.981} & 0.17 & \textbf{3.96} \\
  & all & 0.0060 & 0.042 & \textbf{0.938} & 0.037 & \textbf{0.962} & 0.11 & \textbf{3.68} \\
\hline
 \multirow{4}{*}{Norman holdout} & top 20 & 0.0500 & 0.106 & \textbf{0.991} & 0.102 & \textbf{0.991} & 0.12 & \textbf{3.38} \\
  & top 50 & 0.0200 & 0.061 & \textbf{0.981} & 0.048 & \textbf{0.991} & 0.14 & \textbf{3.72} \\
  & top 100 & 0.0100 & 0.035 & \textbf{0.963} & 0.028 & \textbf{0.981} & 0.17 & \textbf{3.95} \\
  & all & 0.0053 & 0.025 & \textbf{0.947} & 0.022 & \textbf{0.964} & 0.10 & \textbf{3.79} \\
\hline
 \multirow{4}{*}{Replogle RPE1} & top 20 & 0.0500 & 0.155 & \textbf{0.667} & 0.131 & \textbf{0.667} & 0.08 & \textbf{1.55} \\
  & top 50 & 0.0200 & 0.160 & \textbf{0.625} & 0.150 & \textbf{0.600} & 0.29 & \textbf{1.60} \\
  & top 100 & 0.0100 & 0.070 & \textbf{0.605} & 0.070 & \textbf{0.620} & 0.15 & \textbf{1.82} \\
  & all & 0.0010 & 0.017 & \textbf{0.340} & 0.015 & \textbf{0.296} & 0.02 & \textbf{0.64} \\
\hline
\end{tabular}
\end{table*}

\section{Generated populations }
\label{app:umap-density}

\begin{figure}[t]
\centering
\includegraphics[width=\textwidth]{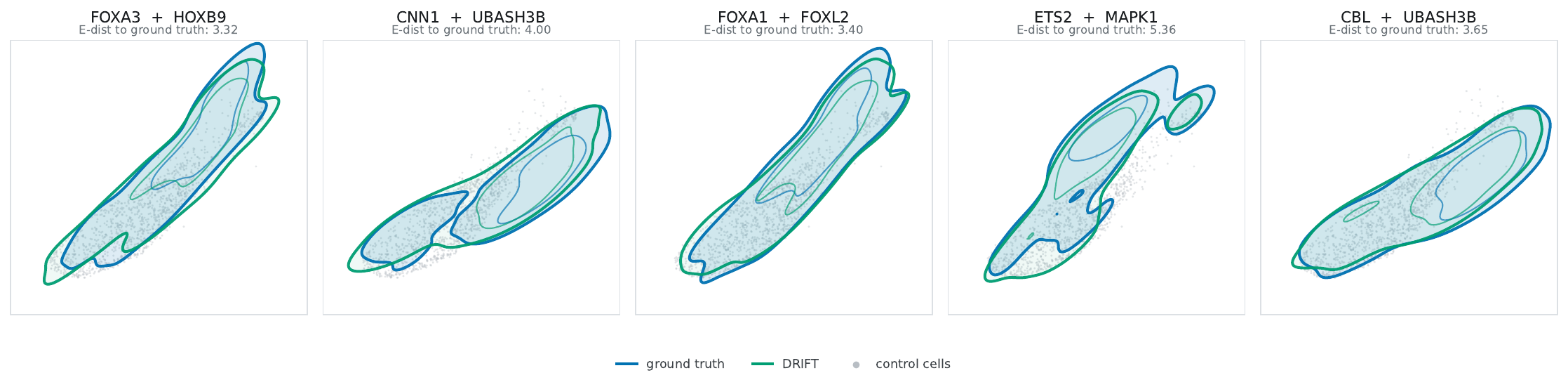}
\caption{UMAP density visualizations on representative Norman additive perturbations.}
\label{fig:umap-density}
\end{figure}

Figure~\ref{fig:umap-density} compares generated and observed cells for the five held-out
conditions of the Norman additive split with the largest shift from control. The generated densities match the observed ones in location and shape,
including the second mode of CNN1+UBASH3B and ETS2+MAPK1. Table~\ref{tab:dist-additive} extends the comparison to all held-out
conditions: DRIFT improves on every baseline on the three distributional metrics, raising
Discr.~Cos from $0.726$ to $0.912$ and reducing E-Dist by $22\%$ relative to scBIG.

\begin{table}[t]
\centering\footnotesize
\caption{Distributional metrics on Norman additive (Appendix~\ref{app:metrics}). Baselines are reported by \citet{scbig2026}; DRIFT is the mean and standard deviation over five seeds. \textbf{Bold}: best; \underline{underline}: second best.}
\label{tab:dist-additive}
\setlength{\tabcolsep}{6pt}\renewcommand{\arraystretch}{1.1}
\begin{tabular}{|c|l||ccc|}
\hline\thickhline
\rowcolor{mygray}
\textbf{Type} & \textbf{Method} & \textbf{Discr. Cos}$\uparrow$ & \textbf{E-Dist}$\downarrow$ & \textbf{Wasserstein}$\downarrow$ \\
\hline\hline
Sta. & Control & -0.3009 & 13.5125 & 16.5615 \\
 & Linear & 0.3967 & 11.1376 & 15.3741 \\
Found. & scFoundation & 0.6218 & 1.3710 & 11.2452 \\
 & GeneCompass & 0.5623 & 1.3059 & 11.1429 \\
Gra. & GEARS & 0.5871 & 1.2377 & 11.0829 \\
Gen. & CellFlow & 0.6397 & 1.1979 & 10.7579 \\
 & scBIG & \underline{0.7264} & \underline{0.8383} & \underline{9.7842} \\
\rowcolor{mygray1}
 & \textbf{DRIFT (ours)} & \textbf{0.9123{\tiny$\pm$0.0039}} & \textbf{0.6516{\tiny$\pm$0.0298}} & \textbf{9.4115{\tiny$\pm$0.0548}} \\
\hline
\end{tabular}
\end{table}

\section{Ablations}
\label{app:ablations}

Table~\ref{tab:ablation-components} removes one component at a time, holding all others and the
training budget fixed. \emph{Without disentanglement} replaces the split latent representation
with a single block trained as a standard variational autoencoder, without a conditional prior or
invariance penalty. The conditional flow consequently transports the full latent representation,
whereas \textsc{Drift} transports only its responsive components, isolating the value of latent
disentanglement. \emph{Without invariance penalty} retains the split latent representation but
sets $\lambda_{\mathrm{CLUB}}=0$. \emph{Without conditioning regularization} removes the mechanism
described in Appendix~\ref{app:condpath} that structures the condition code $e_u$. Because the
metrics have different scales, Table~\ref{tab:ablation-delta-rel} reports each component's
contribution as a fraction of DRIFT's margin over the Linear baseline.
 
\paragraph{Disentanglement.} The disentanglement is the dominant component: it improves all $20$
entries and yields the largest mean contribution on every dataset, ranging from $11\%$ on the
additive split to $35\%$ on holdout double perturbations ($21\%$ overall).

\paragraph{Invariance.} The invariance penalty provides an additional benefit beyond the split
latent. It contributes positively on the majority of  metrics. Its mean contribution is positive on every dataset ($6\%$ overall), with its largest effect on direction of change ($\mathrm{ACC}\Delta$, $17\%$ overall).

\paragraph{Conditioning regularization.} Conditioning regularization improves generalization to unseen
perturbations by structuring the condition representation. Its effect is largest on Replogle ($6.3\%$),
where every test gene is unseen during training: it raises PDS from $0.573$ to $0.604$, accounting for
$43\%$ of DRIFT's margin over Linear. This gain comes with a small $5\%$ decrease in $\rho\Delta$,
but DRIFT still substantially outperforms scBIG. The effect is smallest on the additive split, where
test pairs combine genes observed individually during training.

\paragraph{Perturbation features.} Table~\ref{tab:ablation-features} removes the STRING and
Reactome features from the perturbation context, retaining ESM2 alone. These additional features
are most valuable when the model must generalize to unseen perturbed genes. On Replogle, they
improve all nine metrics. By contrast, their effect is limited on the Norman splits, where test
conditions recombine genes observed individually during training and most metrics are
indistinguishable. Notably, DRIFT with ESM2 alone outperforms \citet{scbig2026} on every split.
Thus, STRING and Reactome features strengthen generalization to unseen genes but do not account
for DRIFT's advantage over scBIG, which also uses these features. Together, these ablations
identify disentanglement as the primary contributor, with invariance and conditioning
regularization providing additional gains.

\begin{table}[t]
\centering\footnotesize
\caption{Component ablations. Each variant removes only the named component. Rank is the mean rank across five metrics (1 = best), averaged over 5 seeds.}
\label{tab:ablation-components}
\setlength{\tabcolsep}{5pt}\renewcommand{\arraystretch}{1.12}
\begin{tabular}{|l||ccccc||c|}
\hline\thickhline
\rowcolor{mygray}
\textbf{Configuration} & $\rho\Delta\uparrow$ & $\rho\Delta^{\mathrm{D}}\uparrow$ & $\mathrm{ACC}\Delta\uparrow$ & DES$\uparrow$ & PDS$\uparrow$ & \textbf{rank} \\
\hline\hline
\multicolumn{7}{|l|}{\emph{Norman additive}} \\
\rowcolor{mygray1}
\textbf{DRIFT} & 0.8871 & 0.8914 & 0.9222 & 0.9017 & 0.9167 & \textbf{1.40} \\
w/o disentanglement & 0.8588 & 0.8236 & 0.9220 & 0.8753 & 0.9042 & 3.40 \\
w/o conditioning regularization & 0.8974 & 0.8926 & 0.9183 & 0.8965 & 0.9051 & 1.80 \\
w/o invariance penalty & 0.8706 & 0.8778 & 0.9146 & 0.8964 & 0.8992 & 3.40 \\
\hline
\multicolumn{7}{|l|}{\emph{Norman holdout (single)}} \\
\rowcolor{mygray1}
\textbf{DRIFT} & 0.6690 & 0.7046 & 0.7890 & 0.6904 & 0.7596 & \textbf{1.40} \\
w/o disentanglement & 0.6378 & 0.6600 & 0.7853 & 0.6598 & 0.7538 & 3.40 \\
w/o conditioning regularization & 0.6662 & 0.7093 & 0.7852 & 0.6811 & 0.7522 & 2.40 \\
w/o invariance penalty & 0.6585 & 0.6970 & 0.7801 & 0.6751 & 0.7606 & 2.80 \\
\hline
\multicolumn{7}{|l|}{\emph{Norman holdout (double)}} \\
\rowcolor{mygray1}
\textbf{DRIFT} & 0.8562 & 0.9319 & 0.8724 & 0.8734 & 0.8075 & \textbf{1.60} \\
w/o disentanglement & 0.7943 & 0.8540 & 0.8571 & 0.8141 & 0.7915 & 4.00 \\
w/o conditioning regularization & 0.8600 & 0.9280 & 0.8709 & 0.8779 & 0.7973 & 2.00 \\
w/o invariance penalty & 0.8534 & 0.9426 & 0.8654 & 0.8657 & 0.8062 & 2.40 \\
\hline
\multicolumn{7}{|l|}{\emph{Replogle RPE1}} \\
\rowcolor{mygray1}
\textbf{DRIFT} & 0.5085 & 0.6160 & 0.6537 & 0.7021 & 0.6044 & \textbf{1.60} \\
w/o disentanglement & 0.4804 & 0.5942 & 0.6452 & 0.6723 & 0.5739 & 3.80 \\
w/o conditioning regularization & 0.5194 & 0.6254 & 0.6569 & 0.6976 & 0.5729 & 1.80 \\
w/o invariance penalty & 0.5055 & 0.6159 & 0.6528 & 0.6931 & 0.5978 & 2.80 \\
\hline
\end{tabular}
\end{table}

\begin{table*}[t]
\centering\footnotesize
\caption{Contribution of genetic perturbation features. \textsc{Drift} represents each perturbed gene using
ESM2\,$\oplus$\,STRING\,$\oplus$\,Reactome; \emph{ESM2 only} removes STRING and Reactome while
leaving all other components unchanged. \citet{scbig2026} already integrates all three sources.
\textbf{Bold}: best; \underline{underline}: second best.}
\label{tab:ablation-features}
\setlength{\tabcolsep}{4pt}\renewcommand{\arraystretch}{1.1}
\resizebox{\textwidth}{!}{%
\begin{tabular}{|l|l||ccccccccc|}
\hline\thickhline
\rowcolor{mygray}
\textbf{Split} & \textbf{Method} & $\rho\Delta\uparrow$ & $\rho\Delta^{\mathrm{D}}\uparrow$ &
$\mathrm{ACC}\Delta\uparrow$ & $\mathrm{ACC}\Delta^{\mathrm{D}}\uparrow$ & DES$\uparrow$ &
PDS$\uparrow$ & L2$\downarrow$ & MSE$\downarrow$ & MAE$\downarrow$ \\
\hline\hline
 \multirow{3}{*}{Norman additive} & scBIG & 0.8496 & 0.8230 & 0.9197 & 0.9906 & 0.8593 & 0.8548 & 3.92 & 0.0091 & 0.0689 \\
  & \textbf{DRIFT}, ESM2 only & \textbf{0.8878} & \textbf{0.8938} & \underline{0.9203} & \textbf{0.9989} & \textbf{0.9020} & \underline{0.9045} & \underline{3.22} & \underline{0.0062} & \underline{0.0569} \\
  & \cellcolor{mygray1}\textbf{DRIFT} & \cellcolor{mygray1}\underline{0.8871} & \cellcolor{mygray1}\underline{0.8914} & \cellcolor{mygray1}\textbf{0.9222} & \cellcolor{mygray1}\underline{0.9977} & \cellcolor{mygray1}\underline{0.9017} & \cellcolor{mygray1}\textbf{0.9167} & \cellcolor{mygray1}\textbf{3.12} & \cellcolor{mygray1}\textbf{0.0057} & \cellcolor{mygray1}\textbf{0.0549} \\
\hline
 \multirow{3}{*}{Replogle RPE1} & scBIG & 0.4875 & 0.5925 & \underline{0.6471} & 0.8089 & 0.5145 & 0.5520 & 6.12 & 0.0118 & 0.0676 \\
  & \textbf{DRIFT}, ESM2 only & \underline{0.4919} & \underline{0.6032} & 0.6465 & \underline{0.8124} & \underline{0.6865} & \underline{0.5696} & \underline{6.04} & \underline{0.0115} & \underline{0.0670} \\
  & \cellcolor{mygray1}\textbf{DRIFT} & \cellcolor{mygray1}\textbf{0.5085} & \cellcolor{mygray1}\textbf{0.6160} & \cellcolor{mygray1}\textbf{0.6537} & \cellcolor{mygray1}\textbf{0.8149} & \cellcolor{mygray1}\textbf{0.7021} & \cellcolor{mygray1}\textbf{0.6044} & \cellcolor{mygray1}\textbf{5.68} & \cellcolor{mygray1}\textbf{0.0101} & \cellcolor{mygray1}\textbf{0.0633} \\
\hline
\end{tabular}}
\end{table*}

\begin{table}[t]
\centering\footnotesize
\caption{Component contributions, expressed as a percentage of DRIFT's margin over the Linear baseline: $(\text{DRIFT}-\text{without component})/(\text{DRIFT}-\text{Linear})$ for each dataset and metric. \textcolor{green!45!black}{Green} values indicate improvement; \textcolor{gray}{grey} values lie within standard error across seeds; \textcolor{red!70!black}{red} values are negative.}
\label{tab:ablation-delta-rel}
\setlength{\tabcolsep}{5pt}\renewcommand{\arraystretch}{1.12}
\begin{tabular}{|l||ccccc||c|}
\hline\thickhline
\rowcolor{mygray}
\textbf{Component} & $\rho\Delta$ & $\rho\Delta^{\mathrm{D}}$ & $\mathrm{ACC}\Delta$ & DES & PDS & \textbf{mean} \\
\hline\hline
\multicolumn{7}{|l|}{\textbf{Norman additive}} \\
Disentanglement & \textcolor{green!45!black}{+10.6} & \textcolor{green!45!black}{+36.9} & \textcolor{green!45!black}{+0.2} & \textcolor{green!45!black}{+4.5} & \textcolor{green!45!black}{+3.3} & +11.1 \\
Conditioning regularization & \textcolor{red!70!black}{$-$3.9} & \textcolor{gray}{$-$0.7} & \textcolor{green!45!black}{+3.2} & \textcolor{green!45!black}{+0.9} & \textcolor{green!45!black}{+3.0} & +0.5 \\
Invariance penalty & \textcolor{green!45!black}{+6.2} & \textcolor{green!45!black}{+7.4} & \textcolor{green!45!black}{+6.3} & \textcolor{green!45!black}{+0.9} & \textcolor{green!45!black}{+4.6} & +5.1 \\
\hline
\multicolumn{7}{|l|}{\textbf{Norman holdout (single)}} \\
Disentanglement & \textcolor{green!45!black}{+40.7} & \textcolor{green!45!black}{+43.5} & \textcolor{green!45!black}{+19.0} & \textcolor{green!45!black}{+13.0} & \textcolor{green!45!black}{+2.3} & +23.7 \\
Conditioning regularization & \textcolor{green!45!black}{+3.7} & \textcolor{gray}{$-$4.6} & \textcolor{green!45!black}{+19.5} & \textcolor{green!45!black}{+3.9} & \textcolor{green!45!black}{+3.0} & +5.1 \\
Invariance penalty & \textcolor{green!45!black}{+13.7} & \textcolor{green!45!black}{+7.4} & \textcolor{green!45!black}{+45.6} & \textcolor{green!45!black}{+6.5} & \textcolor{gray}{$-$0.4} & +14.6 \\
\hline
\multicolumn{7}{|l|}{\textbf{Norman holdout (double)}} \\
Disentanglement & \textcolor{green!45!black}{+44.8} & \textcolor{green!45!black}{+74.8} & \textcolor{green!45!black}{+31.6} & \textcolor{green!45!black}{+19.1} & \textcolor{green!45!black}{+5.2} & +35.1 \\
Conditioning regularization & \textcolor{gray}{$-$2.7} & \textcolor{green!45!black}{+3.7} & \textcolor{green!45!black}{+3.1} & \textcolor{red!70!black}{$-$1.5} & \textcolor{green!45!black}{+3.3} & +1.2 \\
Invariance penalty & \textcolor{green!45!black}{+2.0} & \textcolor{gray}{$-$10.3} & \textcolor{green!45!black}{+14.5} & \textcolor{green!45!black}{+2.5} & \textcolor{green!45!black}{+0.4} & +1.8 \\
\hline
\multicolumn{7}{|l|}{\textbf{Replogle RPE1}} \\
Disentanglement & \textcolor{green!45!black}{+11.9} & \textcolor{green!45!black}{+8.6} & \textcolor{green!45!black}{+11.8} & \textcolor{green!45!black}{+6.1} & \textcolor{green!45!black}{+42.0} & +16.1 \\
Conditioning regularization & \textcolor{red!70!black}{$-$4.6} & \textcolor{red!70!black}{$-$3.7} & \textcolor{red!70!black}{$-$4.4} & \textcolor{green!45!black}{+0.9} & \textcolor{green!45!black}{+43.4} & +6.3 \\
Invariance penalty & \textcolor{green!45!black}{+1.3} & 0.0 & \textcolor{green!45!black}{+1.2} & \textcolor{green!45!black}{+1.8} & \textcolor{green!45!black}{+9.1} & +2.7 \\
\hline
\multicolumn{7}{|l|}{\textbf{All datasets}} \\
Disentanglement & \textcolor{green!45!black}{+27.0} & \textcolor{green!45!black}{+40.9} & \textcolor{green!45!black}{+15.6} & \textcolor{green!45!black}{+10.7} & \textcolor{green!45!black}{+13.2} & +21.5 \\
Conditioning regularization & \textcolor{gray}{$-$1.9} & \textcolor{gray}{$-$1.3} & \textcolor{green!45!black}{+5.3} & \textcolor{green!45!black}{+1.1} & \textcolor{green!45!black}{+13.2} & +3.3 \\
Invariance penalty & \textcolor{green!45!black}{+5.8} & \textcolor{green!45!black}{+1.1} & \textcolor{green!45!black}{+16.9} & \textcolor{green!45!black}{+2.9} & \textcolor{green!45!black}{+3.4} & +6.0 \\
\hline
\end{tabular}
\end{table}

\section{Full results with standard deviations}
\label{app:results-std}

\begin{table*}[t]
\centering\footnotesize
\caption{Standard deviation over five seeds, Norman additive split.}
\label{tab:std-additive}
\setlength{\tabcolsep}{4pt}\renewcommand{\arraystretch}{1.1}
\resizebox{\textwidth}{!}{%
\begin{tabular}{|l||ccccccccc|}
\hline\thickhline
\rowcolor{mygray}
\textbf{Method} & $\rho\Delta$ & $\rho\Delta^{\mathrm{D}}$ & $\mathrm{ACC}\Delta$ & $\mathrm{ACC}\Delta^{\mathrm{D}}$ & DES & PDS & L2 & MSE & MAE \\
\hline\hline
GEARS & 0.0055 & 0.0151 & 0.0031 & 0.0080 & 0.0081 & 0.0048 & 0.0649 & 0.0006 & 0.0014 \\
scGPT & 0.0006 & 0.0045 & 0.0010 & 0.0010 & 0.0084 & 0.0026 & 0.0197 & 0.0002 & 0.0004 \\
scFoundation & 0.0009 & 0.0031 & 0.0004 & 0.0036 & 0.0067 & 0.0029 & 0.0044 & 0.0001 & 0.0002 \\
CellFlow & 0.0035 & 0.0146 & 0.0013 & 0.0213 & 0.0033 & 0.0074 & 0.0636 & 0.0004 & 0.0014 \\
scBIG & 0.0026 & 0.0038 & 0.0023 & 0.0029 & 0.0015 & 0.0096 & 0.0411 & 0.0002 & 0.0009 \\
\hline
\rowcolor{mygray1}
\textbf{DRIFT (ours)} & 0.0034 & 0.0045 & 0.0031 & 0.0019 & 0.0028 & 0.0101 & 0.0558 & 0.0002 & 0.0010 \\
\hline
\end{tabular}}
\end{table*}

\begin{table*}[t]
\centering\footnotesize
\caption{Standard deviation over five seeds, Norman holdout split.}
\label{tab:std-holdout}
\setlength{\tabcolsep}{4pt}\renewcommand{\arraystretch}{1.1}
\resizebox{\textwidth}{!}{%
\begin{tabular}{|l||ccccccccc|}
\hline\thickhline
\rowcolor{mygray}
\textbf{Method} & $\rho\Delta$ & $\rho\Delta^{\mathrm{D}}$ & $\mathrm{ACC}\Delta$ & $\mathrm{ACC}\Delta^{\mathrm{D}}$ & DES & PDS & L2 & MSE & MAE \\
\hline\hline
scGPT & 0.0008 & 0.0027 & 0.0003 & 0.0031 & 0.0155 & 0.0006 & 0.0039 & 0.0001 & 0.0001 \\
scFoundation & 0.0072 & 0.0122 & 0.0042 & 0.0091 & 0.0158 & 0.0023 & 0.0227 & 0.0000 & 0.0005 \\
CellFlow & 0.0026 & 0.0098 & 0.0013 & 0.0038 & 0.0042 & 0.0092 & 0.0053 & 0.0000 & 0.0001 \\
scBIG & 0.0016 & 0.0036 & 0.0021 & 0.0014 & 0.0042 & 0.0026 & 0.0105 & 0.0000 & 0.0002 \\
\hline
\rowcolor{mygray1}
\textbf{DRIFT (ours)}& 0.0048 & 0.0041 & 0.0023 & 0.0046 & 0.0061 & 0.0040 & 0.0238 & 0.0001 & 0.0004 \\
\hline
\end{tabular}}
\end{table*}

\begin{table*}[t]
\centering\footnotesize
\caption{Replogle RPE1, mean and standard deviation over 5 seeds of \textsc{DRIFT}.}
\label{tab:replogle-std}
\setlength{\tabcolsep}{4pt}\renewcommand{\arraystretch}{1.1}
\resizebox{\textwidth}{!}{%
\begin{tabular}{|c|l||ccccccccc|}
\hline\thickhline
\rowcolor{mygray}
\textbf{Type} & \textbf{Method} & $\rho\Delta\uparrow$ & $\rho\Delta^{\mathrm{D}}\uparrow$ & $\mathrm{ACC}\Delta\uparrow$ & $\mathrm{ACC}\Delta^{\mathrm{D}}\uparrow$ & DES$\uparrow$ & PDS$\uparrow$ & L2$\downarrow$ & MSE$\downarrow$ & MAE$\downarrow$ \\
\hline\hline
\rowcolor{mygray1}
\rotatebox{90}{Gen.} & \textbf{DRIFT (ours)} & 0.5085{\tiny$\pm$0.0074} & 0.6160{\tiny$\pm$0.0058} & 0.6537{\tiny$\pm$0.0013} & 0.8149{\tiny$\pm$0.0015} & 0.7021{\tiny$\pm$0.0072} & 0.6044{\tiny$\pm$0.0067} & 5.68{\tiny$\pm$0.03} & 0.0101{\tiny$\pm$0.0001} & 0.0633{\tiny$\pm$0.0002} \\
\hline
\end{tabular}}
\end{table*}

\begin{table*}[t]
\centering\footnotesize
\caption{ComboSciPlex, mean and standard deviation over five end-to-end seeds of \textsc{Drift}.}
\label{tab:combo-std}
\setlength{\tabcolsep}{6pt}\renewcommand{\arraystretch}{1.1}
\resizebox{\textwidth}{!}{%
\begin{tabular}{|l||cccccc|}
\hline\thickhline
\rowcolor{mygray}
\textbf{Method} & $\rho\Delta\uparrow$ & DE-Spearman $\rho\uparrow$ & DS$\uparrow$ &
L2$\downarrow$ & MSE$\downarrow$ & MAE$\downarrow$ \\
\hline\hline
\rowcolor{mygray1}
\textbf{DRIFT (ours)} & 0.9467{\tiny$\pm$0.0057} & 0.8901{\tiny$\pm$0.0057} & 0.8816{\tiny$\pm$0.0171} & 1.2804{\tiny$\pm$0.0536} & 0.0019{\tiny$\pm$0.0002} & 0.0150{\tiny$\pm$0.0006} \\
\hline
\end{tabular}}
\end{table*}

\end{document}